\documentclass[10pt,journal,compsoc]{IEEEtran}
\ifCLASSOPTIONcompsoc
  \usepackage[nocompress]{cite}
\else
  \usepackage{cite}
\fi
\ifCLASSINFOpdf
\else
\fi
\usepackage{graphics} 
\usepackage{epsfig} 
\usepackage{bbding} 
\usepackage{amsmath}
\DeclareMathOperator*{\argmax}{arg\,max}

\usepackage{bm}
\newcommand\numberthis{\addtocounter{equation}{1}\tag{\theequation}}

\usepackage{wrapfig}
\usepackage{multirow}

\usepackage{subcaption}
\usepackage[labelformat=parens,labelsep=quad,skip=3pt]{caption}
\usepackage{graphicx}

\usepackage{adjustbox}
\usepackage{color,xcolor,colortbl}
\usepackage{amsfonts}

\begin{document}
%
\title{Improving Online Forums Summarization via Hierarchical Unified Deep Neural Network}
%
%
%
%

\author{Sansiri~Tarnpradab,~\IEEEmembership{Member,~IEEE},
        Fereshteh Jafariakinabad,~\IEEEmembership{Member,~IEEE},\\
        and~Kien A.~Hua,~\IEEEmembership{~Fellow,~IEEE}
\IEEEcompsocitemizethanks{\IEEEcompsocthanksitem Sansiri Tarnpradab, 
Fereshteh Jafariakinabad, and Kien A. Hua are with the Department
of Computer Science, University of Central Florida, Orlando, FL, 32816. \protect
E-mail: sansiri@knights.ucf.edu, fereshteh.jafari@knights.ucf.edu, kienhua@cs.ucf.edu}
\thanks{Manuscript received April 19, 2005; revised August 26, 2015.}}

%
%

\markboth{Journal of \LaTeX\ Class Files,~Vol.~14, No.~8, August~2015}%
{Shell \MakeLowercase{\textit{et al.}}: Bare Demo of IEEEtran.cls for Computer Society Journals}
%



\IEEEtitleabstractindextext{%
\begin{abstract}
Online discussion forums are prevalent and easily accessible, thus allowing people to share ideas and opinions by posting messages in the discussion threads. Forum threads that significantly grow in length can become difficult for participants, both newcomers and existing, to grasp main ideas. To mitigate this problem, this study aims to create an automatic text summarizer for online forums. We present Hierarchical Unified Deep Neural Network to build sentence and thread representations for the forum summarization. In this scheme, Bi-LSTM derives a representation that comprises information of the whole sentence and whole thread; whereas, CNN captures most informative features with respect to context from sentence and thread. Attention mechanism is applied on top of CNN to further highlight high-level representations that carry important information contributing to a desirable summary. Extensive performance evaluation has been conducted on three datasets, two of which are real-life online forums and one is news dataset. The results reveal that the proposed model outperforms several competitive baselines.
\end{abstract}
}

\maketitle

\IEEEdisplaynontitleabstractindextext

%
\IEEEpeerreviewmaketitle

\IEEEraisesectionheading{\section{Introduction}\label{sec:intro}}

%
%
%
%

\IEEEPARstart{O}{nline} discussion forums embody a plethora of information exchanged among people with a common interest. Typically, a discussion thread is initiated by a user posting a message (e.g. question, suggestion, narrative, etc), then other users who are interested in the topic will join the discussion, also by posting their own messages (e.g. answer, relevant experience, new question, etc). The thread that gains popularity can span hundreds of messages, putting burden on both newcomers and current participants as they have to spend extra time to understand or simply to catch up with the discussion so far.  An automatic forum summarization method that generates a concise summary is therefore highly desirable.

One simple way to produce a summary is to identify salient sentences and aggregate them. This method naturally aligns with the concept of extractive summarization which involves selecting representative units and concatenating them according to their chronological order.  In order to determine saliency of each unit, the \textit{context} must be taken into account. This factor is critical to any summarization process whether it be an automatic system or a human tasked with selecting sentences from a document to form a summary. As an illustration, if a human is given a thread to extract key sentences from, he/she would first read the thread to grasp contextual information, then select sentences based on that context to compose a summary. On the other hand, if an arbitrary sentence is shown to a human without supplying context of the thread which that sentence belongs to, there would be no clear way of deciding if the sentence should be included in the summary. Previous works  \cite{cheng2016neural, yang-etal-2016-hierarchical, zhou-etal-2018-neural-document} have shown that a context information lies within a document structure contributes to a performance improvement of a summarizer. Similar to documents, forum threads also possess a hierarchical structure; in which, words constitute a sentence, sentences constitute a post, and posts constitute a thread. 

In this work, we propose a data-driven hierarchical-based approach to summarize online forums. In order to utilize knowledge of the forum structure, the method hierarchically encodes sentences and threads to obtain sentence and thread representations. Meanwhile, an attention mechanism is applied to further place emphasis on salient units. Drawing our inspiration from how humans read, comprehend, and then summarize a document, it led us to a network design that unifies Bidirectional Long Short-Term Memory (Bi-LSTM) and Convolutional Neural Network (CNN). 
In this scheme, Bi-LSTM derives a representation that comprises information of an entire sentence and thread; whereas, CNN captures most informative features.
All in all, both networks are utilized with an aim to leverage their individual strength to achieve effective representations, compared to when either one is used. Our extensive experimental results verifies this effectiveness.

The contributions of this study are as follows:
\begin{itemize}
    \item We propose a hierarchical-based unified neural network\footnote{Our code is available at https://github.com/sansiri20/forums\_summ.git} which utilizes Bi-LSTM and CNN to obtain representations for summarizing forum threads. The attention mechanism is employed to give weight to important units. Different from previous studies that apply attention directly to individual words and sentences \cite{yang-etal-2016-hierarchical, tarnpradab2017toward}, our findings suggest that applying attention to the high-level features extracted by CNN contributes to improvements in the performance. 
    \item To demonstrate the advantage of the proposed model, we perform comprehensive empirical study. The results show that the proposed approach significantly outperforms a range of competitive baselines as well as the initial study \cite{tarnpradab2017toward}. This encourages further investigation into the use of the proposed network for text summarization.
    \item We conduct an extensive experiment using different pretrained embeddings (static and contextual) to investigate their effectiveness towards improving the summarization performance. 
\end{itemize}

The remainder of this paper is organized as follows. We review the literature related to automatic summarization in Section \ref{sec:related_works}. The proposed framework is introduced in Section \ref{sec:method}. In Section \ref{sec:experiment}, we provide details on the dataset and the experimental configurations for the performance studies, and explain the baselines used in the comparative study to assess the effectiveness of our proposed model. The performance results are analyzed in Section \ref{sec:results}.  Finally, we draw our conclusions in Section \ref{sec:conclusion}.

\section{Related Work}\label{sec:related_works}
In this study, we address the problem of online forums summarization. Therefore, described herein this section are three major strands of research related to this study, including extractive summarization, neural network-based text summarization, and representation learning.
\subsection{Extractive Summarization}
There are mainly two kinds of methods used in text summarization, namely \textit{extractive summarization} and \textit{abstractive summarization} \cite{hahn2000challenges}. Owing to its effectiveness and simplicity, the extractive summarization approach has been used extensively. The technique involves segmenting text into units (e.g. sentences, phrases, paragraphs, etc), then concatenating a key subset of these units to derive a final summary. In contrast, the abstractive approach functions similarly to paraphrasing, by which the original units are hardly preserved in the output summary. In this study, we consider the extractive summarization approach and propose a deep classifier to recognize key sentences for the summary.

The extractive approach has been applied to data from various domains such as forum threads \cite{bhatia-etal-2014-summarizing, tarnpradab2017toward}, online reviews \cite{hu2017opinion, 10.1145/3357384.3358161, 10.1145/3368926.3369699, di2017surf}, 
emails \cite{carenini-etal-2008-summarizing}, group chats \cite{10.1145/3274465, 10.1145/3159652.3160588}, meetings \cite{10.1145/3279981.3279987, 10.1145/2993148.2993160}, microblogs \cite{rudra2018identifying, 10.1145/3178541, 10.1145/3357384.3358020, zhou2016cminer}, and news \cite{10.1145/3377407, 10.1145/2980258.2980442, 10.1145/3289600.3291008, sethi2017automatic} just to name a few. In the news domain, articles typically follow a clear pattern where the most important point is at the top of the article, followed by the secondary point, and so forth. We generally do not observe a clear pattern in other domains. For example, a forum thread is participated and written by multiple users; thus, the gist may be contained across different posts -- not necessarily at the first sentence or paragraph. Furthermore, these user-generated content (UGC) generally contains noise, misspellings, and informal abbreviations which make choosing sentences for summarization more challenging. In our work, we focus on summarizing content in the forum thread. Given the nature of forum data, it can be framed as a multi-document summarization where these documents are created and posted by different authors.

\subsection{Neural Network-based Text Summarization}\label{sec:related_NN}
A large body of research applies neural networks involving RNN \cite{nallapati2017summarunner}, CNN \cite{cao2017improving}, along with a combination of both \cite{10.1145/3132847.3133127, narayan-etal-2018-ranking} to improve text summarization. 
For example, Nallapati et al.\cite{nallapati2017summarunner} have proposed an RNN-based sequence model entitled SummaRuNNer to produce extractive summaries. A two-layer bidirectional Gated Recurrent Unit (GRU) is applied to derive document representations. The first layer runs at the word level to derive hidden representation of each word in both forward and backward directions. The second layer runs at the sentence level to encode the representations of sentences in the document.
Cao et al.\cite{cao2017improving} have proposed a CNN-based summarization system entitled TCSum to perform multi-document summarization. Adopting transfer learning concept, TCSum demonstrated that the distributed representation projected from text classification model can be shared with the summarization model. The model can achieve state-of-the-art performance without handcrafted features needed. 

A unified architecture that combines RNN and CNN for summarization task has shown success in several works.
For instance, Singh et al.\cite{10.1145/3132847.3133127} have proposed Hybrid MemNet, a data-driven end-to-end network for a single-document summarization where CNN is applied to capture latent semantic features and LSTM is applied thereafter to capture an overall representation of the document. The final document representation is generated by concatenating two document embeddings, one from CNN-LSTM and the other from the memory network.
Narayan et al.\cite{narayan-etal-2018-ranking} also proposed a unified architecture which frames an extractive summarization problem with a reinforcement learning objective. The architecture involves LSTM and CNN to encode sentences and documents successively. The model learns to rank sentences by training the network in a reinforcement learning framework while optimizing ROUGE evaluation metric.

Several lines of research have taken into account the hierarchical structure of the document \cite{10.1145/3308558.3313619, 10.1145/3308558.3313707, 10.1145/3341105.3374025, cheng2016neural, zhou-etal-2018-neural-document}. 
Cheng and Lapata\cite{cheng2016neural} have developed a framework containing a hierarchical document encoder and an attention-based extractor for single-document summarization. The hierarchical information has shown to help derive a meaningful representation of a document. 
Zhou et al.\cite{zhou-etal-2018-neural-document} have proposed an end-to-end neural network framework to generate extractive document summaries. Essentially, the authors have developed a hierarchical encoder via bidirectional Gated Recurrent Unit (BiGRU) which integrates sentence selection strategy into the scoring model, so that the model can jointly learn to score and select sentences.

The usage of an attention mechanism has also proven successful in many applications \cite{rush-etal-2015-neural, 10.1145/3341105.3373892, nema2017diversity, wang-ling-2016-neural, cao-etal-2016-attsum, narayan-etal-2018-document, 10.1145/3269206.3269251}. 
For example, Wang and Ling\cite{wang-ling-2016-neural} have introduced an attention-based encoder-decoder concept to summarize opinions. The authors have applied LSTM network to generate abstracts, where a latent representation computed from the attention-based encoder is an input to the network.
Cao et al.\cite{cao-etal-2016-attsum} have applied the attention concept to simulate human attentive reading behavior for extractive query-focused summarization. The system called AttSum is proposed and demonstrated to be capable of handling query relevance ranking and sentence saliency ranking jointly. 


\subsection{Representation Learning}
Representation learning which aims to acquire representations automatically from the data plays a crucial role in many Natural Language Understanding (NLU) and Natural Language Processing (NLP) models. Particularly, pre-trained word representations are the building blocks of any NLP and NLU models that have shown to improve downstream tasks in many domains such as text classification, machine translation, machine comprehension, among others \cite{camacho2018word}. Learning high-quality word representations is challenging, and many approaches have been developed to produce pre-trained word embeddings which differ on how they model the semantics and context of the words. word2vec \cite{mikolov2013distributed}, a window-based model, and GloVe (\textbf{Glo}bal \textbf{Ve}ctors for Word Representation) \cite{pennington-etal-2014-glove}, a count-based model, rely on distributional language hypothesis in order to capture the semantics. FastText \cite{bojanowski2017enriching} is a character-based word representation in which a word is represented as a bag of character n-grams and the final word vector is the sum of these representations. One of the advantages of FastText is the capability of handling out-of-vocabulary words (OOV) -- unlike word2vec and GloVe. 

Although the classical word embeddings can capture semantic and syntactic characteristics of words to some extent, they fail to capture polysemy and disregard the context in which the word appears. To address the polysemous and context-dependent nature of words, the contextualized word embeddings are proposed. ELMo (\textbf{E}mbeddings from \textbf{L}anguage \textbf{Mo}dels) proposes a deep contextualized word representation in which each representation is a function of the input sentence where the objective function is a bidirectional Language Model (biLM) \cite{peters-etal-2018-deep}. The representations are a linear combination of all of the internal layers of the biLM where the weights are learnable for a specific task. BERT (\textbf{B}idirectional \textbf{E}ncoder \textbf{R}epresentations from \textbf{T}ransformers) is another contextualized word representation which is trained on bidirectional transformers by jointly conditioning on both left and right context in all layers \cite{devlin-etal-2019-bert}. The objective function in BERT is a masked language model where some of the words in the input sentence are randomly masked. FLAIR is contextualized character-level word embedding which models words and context as sequences of characters and is trained on a character-level language model objective \cite{akbik-etal-2018-contextual}. 

In summary, there are several approaches adopted to learn the word representation in literature which differ in the ways they model meaning and context. The choice of word embeddings for particular NLP tasks is still a matter of experimentation and evaluation. In this study, we experimented with word2vec, FastText, ELMo, and BERT embeddings, by integrating them in an embedding layer of the model. These embeddings initialize vectors of words/sentences present in the forum data.

\begin{figure*}
\includegraphics[width=\textwidth]{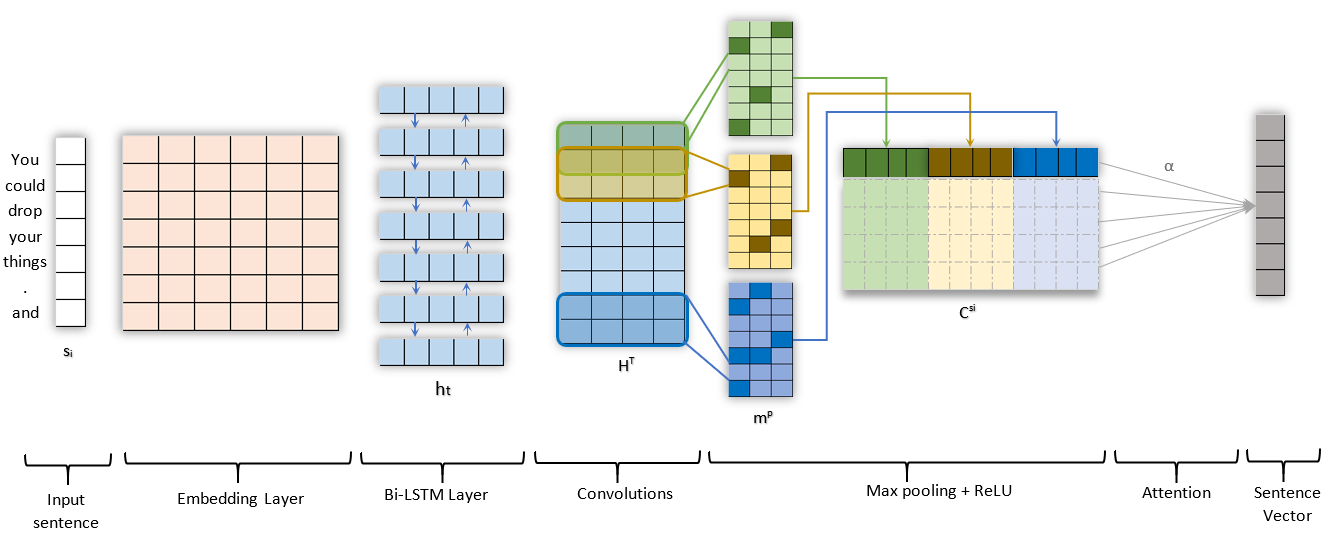} 
\caption{Illustration of the sentence encoder. }
\label{fig:model_arch}
\end{figure*} 

\section{Summarization Model}\label{sec:method}
Our system is tasked with extracting representative sentences from a thread to form a summary, which is naturally well-suited to be formulated as a supervised-learning task.
We consider a sentence as an extraction unit due to its succinctness. Let $\bm{s}$=$[s_1,\cdots,s_N]$ be the sentences in a thread and $\bm{l}$=$[l_1,\cdots,l_N]$ be the corresponding labels, where ``$l_i$ = 1" indicates that the sentence $s_i$ is part of the summary, and ``0" otherwise. Our goal is to find the most probable tag sequence given the thread sentences:
\begin{align*}
    \argmax_{\bm{l} \in \mathcal{T}} p(\bm{l}|\bm{s}) \label{eqn_argmax}\numberthis
\end{align*}
\noindent where $\mathcal{T}$ is the set of all possible tag sequences, and $\textstyle p(\bm{l}|\bm{s}) = \prod_{i=1}^N p(l_i|\bm{s})$ where the tag of each sentence is determined independently.

In this section, we elaborate our hierarchical-based framework for multi-document summarization. 
Inspired by the development of Hierarchical Attention Networks (HAN) \cite{yang-etal-2016-hierarchical, tarnpradab2017toward}, the proposed model adopts their concept to construct sentence and thread representations.
Two types of neural networks, namely bi-directional recurrent neural network and convolutional neural network, are utilized to maximize capability of the summarizer. In a nutshell, the model is comprised of hierarchical encoders, a neural attention component, and a sentence extractor. The encoders generate the representations based on words and sentences in the forum. The neural attention mechanism pinpoints any meaningful units in the process. Finally, the sentence extractor selects and puts together all the key sentences to produce a summary. In the following, the boldface letters represent vectors and matrices. Words and sentences are denoted by their indices.


\subsection{Sentence Encoder}
The sentence encoder reads an input sentence as a sequence of word embeddings, then returns a sentence vector as an output. Adopting the pipeline architecture to process data in a streaming manner, a bi-directional recurrent neural network is followed by a convolutional neural network to constitute the sentence encoder. Furthermore, the attention mechanism is employed while generating the sentence vector to give more emphasis on units that contribute more to the meaning of the sentence. This strategy to sentence encoding is illustrated in Figure \ref{fig:model_arch}. We elaborate the different network components in the following subsections. 

\subsubsection{Input Layer}
Given that, each thread is a sequence of sentences and each sentence is a sequence of words, let $\bm{s}_i$=$[x_1,\cdots,x_T]$ denote the $i$-th sentence and the words are indexed by $t \in {\{1,\cdots,T\}}$ where $T$ denotes the number of words in the sentence.
Each word is converted to its corresponding pretrained embedding, and subsequently fed into the bidirectional recurrent neural network. 

\subsubsection{Bidirectional Recurrent Neural Network layer}
We opt for Bidirectional Long Short-Term Memory (Bi-LSTM) due to its effectiveness as evidenced in previous studies \cite{hochreiter1997long}. LSTM contains an input gate ($i_t$), a forget gate ($f_t$), and an output gate ($o_t$) to control the amount of information coming from the previous time-step as well as flowing out in the next time-step. This gating mechanism accommodates long-term dependencies by allowing the information flow to sustain for a long period of time. Our Bi-LSTM model contains forward pass and backward pass (Eq. \ref{forward_h}-\ref{backward_h}). 
The forward hidden representation $\overrightarrow{\bm{h}_t}$ comprises semantic information from the beginning of the sentence to the current time-step; on the contrary,  $\overleftarrow{\bm{h}_t}$ comprises semantic information from the current time-step to the end of the sentence. 
Both vectors $\overrightarrow{\bm{h}_t}$ and  $\overleftarrow{\bm{h}_t}$ are of dimension $\mathbb{R}^{d_t}$, where $d_t$ is the dimensionality of the hidden state in the word-level Bi-LSTM. 
Finally, concatenating the two vectors,  $\bm{h}_t$=$[\overrightarrow{\bm{h}_t}, \overleftarrow{\bm{h}_t}] \in \mathbb{R}^{2d_t}$, produces a word representation that carries contextual information of the whole sentence the word being a part of.
\begin{align*}
\overrightarrow{\bm{h}_t}=\text{LSTM}_1(\overrightarrow{\bm{h}_{t-1}}, \bm{x}_t) \label{forward_h} \numberthis\\
\overleftarrow{\bm{h}_t}=\text{LSTM}_2(\overleftarrow{\bm{h}_{t-1}}, \bm{x}_t) \label{backward_h} \numberthis
\end{align*}


\subsubsection{Convolutional Layer}
The convolutional layer is primarily used to capture most informative features. The representation $\bm{h}_t$ of every word in the sentence $\bm{s}_i$ are compiled to form a matrix $\bm{H}^{s_i}$ which is an input to the CNN. Concretely, $\bm{H}^{s_i}$ = $[ \bm{h}_1,\cdots,\bm{h}_T ]$, where $\bm{H}^{s_i} \in \mathbb{R}^{T\times2d_t}$. The convolutional layer is composed of a set of filters $\bm{F}$. 
Each filter $\bm{F}^p \in \mathbb{R}^{j\times2d_t}$, where $p$ denotes filter index, slides across the input with a window of $j$ words to form a feature map $\bm{m^p}\in \mathbb{R}^{T-j+1}$.  
\begin{align*}
\bm{m}^{p} = ReLU(\bm{F}^p\cdot\bm{H}^{s_i}_{[a:a+j-1]}+b), a \in [1, T-j+1 ] \label{m_feature_map} \numberthis
\end{align*}
\noindent where 
$\bm{H}^{s_i}_{[a:a+j-1]}$ denotes a submatrix of ${H}^{s_i}$ from row $a$ to row $a+j-1$;
$b \in \mathbb{R}$ is an additive bias. A \textit{Rectified Linear Unit} (ReLU) is applied element-wise as a nonlinear activation function in our study. 

One-dimensional max-pooling operation is then performed to obtain a fixed-length vector. 
Given that each feature $\bm{m}^{p}$ is of length $T-j+1$, a feature map $\bm{m}^{p}$  is transformed into a vector of half the length through a 1D max-pooling window of size $2$. 
In other words, only meaningful features \textit{per bigram} are extracted (Eq. \ref{maxpool}). 
All resultant feature maps are combined into a final representation $\bm{C}^{s_i}\in \mathbb{R}^{\lfloor(T-j+1)/2\rfloor\times|F|}$ (Eq. \ref{concat_maxpool}).  
\begin{align*}
\bm{f}^p = [max(\bm{m}_{[1:2]}^p) \oplus \cdots \oplus max(\bm{m}_{[T-j : T-j +1 ]}^p)] \label{maxpool} \numberthis
\end{align*}
\begin{align*}
\bm{C}^{s_i} &= [\bm{f}^{1}\oplus \cdots \oplus \bm{f}^{|F|}] \label{concat_maxpool} \numberthis
\end{align*}

\begin{figure*}
\includegraphics[width=\textwidth]{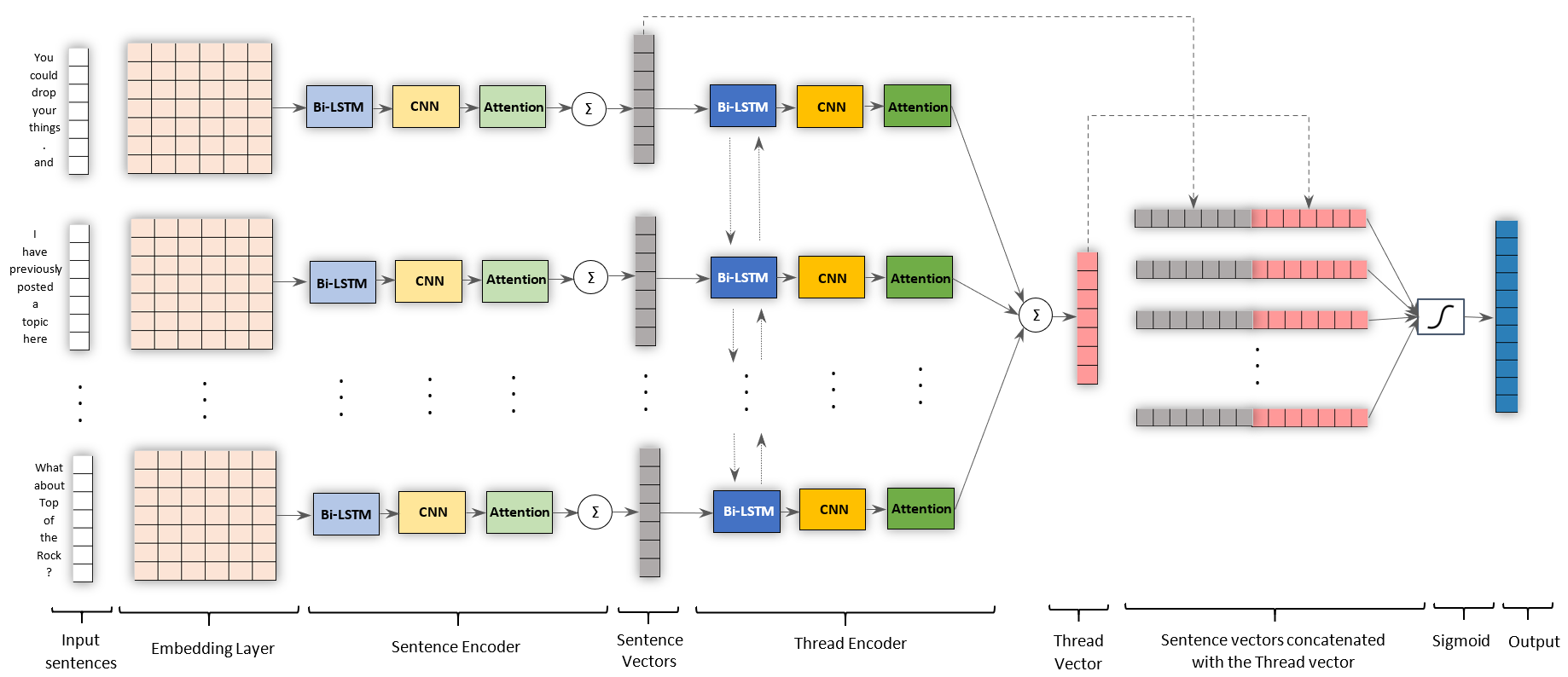} 
\caption{Complete framework of the proposed summarization model.}
\label{fig:full_arch}
\end{figure*}

\subsubsection{Attention Layer}
In this section, we describe the attention mechanism employed to attend to important units (bigrams) in a sentence. 
We introduce a trainable vector $\bm{u}_v$ for all the bigrams to capture ``global" bigram saliency. Each vector of $\bm{C}^{s_i}$, denoted as $\bm{C}^{s_i}_v$, is selected through a multiplication operation $\bm{C}^{{s}_i}e_v$ where $e_v$ is a standard basis vector containing all zeros except for a one in the $v$-th position. 
$\bm{C}^{s_i}_v$ is projected to a transformed space to generate $\bm{u}_{{s}_i}$ (Eq. \ref{u_t}). The inner product $\bm{u}_{{s}_i}^T \bm{u}_v$ signals the importance of the $v$-th bigram. We convert it to a normalized weight $\alpha_{{s}_i}$ using a softmax function (Eq. \ref{alpha_t}). Finally, a weighted sum of representation is computed to obtain a sentence vector $\bm{s}_i$ (Eq. \ref{s_i}).
\begin{align*}
\bm{u}_{{s}_i} &= \tanh(\bm{W}^{{s}_i} \bm{C}^{s_i}_v+\bm{b}^{{s}_i})\label{u_t} \numberthis \\
\alpha_{{s}_i} &= \frac{\exp(\bm{u}_{{s}_i}^T \bm{u}_v)}{\sum_{v} \exp(\bm{u}_{{s}_i}^T \bm{u}_v)} \label{alpha_t}, 1\leq v\leq|F|  \numberthis \\
\bm{s}_i &= \sum_{v} \alpha_{{s}_i} \bm{C}^{s_i}_v\label{s_i} \numberthis
\end{align*}

\subsection{Thread Encoder}
The thread encoder takes as input a sequence of sentence vectors $\bm{d}$=$[\bm{s}_1,\cdots,\bm{s}_N]$ previously encoded through the sentence encoder. We choose to index sentences by $i$. The thread encoder has a similar network architecture as the sentence encoder, summarized by Eq. \ref{h_i_fwd} - \ref{thread_vec}. 
Vectors $\overrightarrow{\bm{h}_i}$ and  $\overleftarrow{\bm{h}_i}$ are of dimension $\mathbb{R}^{d_i}$; thus, $\bm{h}_i=[\overrightarrow{\bm{h}_{i}} , \overleftarrow{\bm{h}_{i}}]\in\mathbb{R}^{2d_i}$ where $d_i$ is the dimensionality of the hidden state in the sentence-level Bi-LSTM (Eq. \ref{h_i_fwd}-\ref{h_i}). 
A matrix $\bm{H}^d\in\mathbb{R}^{N\times2d_i}$ is generated from $\bm{h}_i$ of every sentence in the thread compiled together (Eq. \ref{H_S}). 
\begin{align*}
\overrightarrow{\bm{h}_i} &= \text{LSTM}_3(\overrightarrow{\bm{h}_{i-1}},\bm{s}_{i})\label{h_i_fwd} \numberthis\\
\overleftarrow{\bm{h}_i} &= \text{LSTM}_4(\overleftarrow{\bm{h}_{i-1}}, \bm{s}_{i})\label{h_i_bwd} \numberthis\\
\bm{h}_{i} &= [\overrightarrow{\bm{h}_{i}} , \overleftarrow{\bm{h}_{i}}]\label{h_i} \numberthis\\
\bm{H}^{d} &= [ \bm{h}_1,\cdots,\bm{h}_N] \label{H_S} \numberthis 
\end{align*}
Each feature map is represented by $\bm{m}^{q}\in \mathbb{R}^{N-k+1}$, where $q$ is a filter index; $N$ is total number of sentences in the thread; and $k$ is a filter height (Eq. \ref{m_q}). $\bm{f}^{q}$ is constituted of the max-pooled values of $\bm{m}^{q}$ concatenated together (Eq. \ref{f_q}). The max-pooling window size is 2 representing \textit{a pair of consecutive sentences}. All resultant max-pooled vectors are combined into a final representation $\bm{C}^{d}\in \mathbb{R}^{\lfloor(N-k+1)/2\rfloor\times|F|}$ (Eq. \ref{C_S}). 
\begin{align*}
\bm{m}^{q} = ReLU(\bm{F}^q\cdot\bm{H}^d_{[a:a+k-1]}+b) \label{m_q}, a \in [1, N-k+1]  \numberthis 
\end{align*}
\begin{align*}
\bm{f}^q = [max(\bm{m}_{[1:2]}^q) \oplus \cdots \oplus max(\bm{m}_{[N-k : N-k +1 ]}^q)] \label{f_q}  \numberthis
\end{align*}
\begin{align*}
\bm{C}^d &= [\bm{f}^{1}\oplus \cdots \oplus \bm{f}^{|F|}]\label{C_S} \numberthis 
\end{align*}

Each vector of $\bm{C}^{d}$ is denoted by $\bm{C}^{d}_{v'}$ (Eq. \ref{u_i}).
The sentence-level attention mechanism introduces a \textit{trainable} vector $\bm{u}_{v'}$ that encodes salient sentence-level content. The thread vector $\bm{d}$ is a weighted sum of sentence pairs, where $\alpha_d$ is a normalized scalar value indicating important sentence pairs in the thread (Eq. \ref{alpha_d}, \ref{thread_vec}).
\begin{align*}
\bm{u}_{d} = \tanh(\bm{W}^d \bm{C}^{d}_{v'} + \bm{b}^d);\label{u_i} \numberthis
\end{align*}
\begin{align*}
\alpha_{d} = \frac{\exp(\bm{u}_{d}^T \bm{u}_{v'})}{\sum_{v'} \exp(\bm{u}_{d}^T \bm{u}_{v'})}, 1\leq v'\leq|F|\label{alpha_d}\numberthis
\end{align*}
\begin{align*}
\bm{d} = \sum_{v'} \alpha_{d} \bm{C}^{d}_{v'}\label{thread_vec}\numberthis
\end{align*}

\subsection{Output Layer}
The vector representation of each sentence is concatenated with its corresponding thread representation to construct the final representation, denoted as $[\bm{s}_i,\bm{d}]$. With this, both sentence-level and thread-level context are taken into account for the classification. The learned vector representations are fed into a dense layer of which sigmoid is used as an activation function. Cross-entropy is used to measure the network loss. 

\subsection{Sentence Extraction}
We impose a limit to the number of words in the final summary -- at most 20\% of total words in the original thread are allowed. In order to extract salient sentences, all sentences are sorted based on saliency scores outputted from the dense layer. Sorted sentences are then iteratively added to the final summary until the compression limit is reached. At last, all the sentences in the final summary are chronologically ordered according to their appearance in the original thread. Since sentences selected by supervised summarization models tend to be redundant \cite{cao2017improving}, we apply an additional constraint to include a sentence in the summary only if it contains at least 50\% new bigrams in comparison to all existing bigrams in the final summary.
Henceforth, we refer to our approach as \textit{Hierarchical Unified Deep Neural Network}. A complete framework of the proposed network is illustrated in Figure \ref{fig:full_arch}. 

\section{Experiment}\label{sec:experiment}
In this section, we first give a description of the datasets used for experiments, followed by details of how the training set is created. Then, we present experiment configurations along with a list of hyperparameters explored to achieve the best performing model. Next, we provide a brief description of baselines used in our performance study, and subsequently give an introduction to the metrics for evaluating performance of the summarization system. 
\subsection{Dataset}
Since the proposed approach is applicable for multi-document summarization, we perform experiments on news articles in addition to online forums. Three datasets, namely Trip advisor, Reddit, and Newsroom, are used in our study -- the former two were crawled from online forums while the other is news articles from major publications. Statistics of all datasets is provided in Table \ref{tab:stats} and a brief description of each is as follows.

\textbf{Trip advisor.} The Trip advisor\footnote{https://www.tripadvisor.com/} forum data were collected by Bhatia et al.\cite{bhatia-etal-2014-summarizing}. In our study, there are a total of 700 TripAdvisor threads, 100 of which were originally annotated with human summaries by \cite{bhatia-etal-2014-summarizing}, and the additional 600 threads were annotated later by \cite{tarnpradab2017toward}. We held out 100 threads as a development set and reported the performance results on the remaining threads. The development set is mainly used for hyperparameter-tuning purposes as described in Section \ref{model_config}. The reference summaries were prepared by having two human annotators generate a summary for each thread. Both annotators were instructed to read a thread, then write a corresponding summary with the length limited within 10\% to 25\% of the original thread length. The annotators were also encouraged to pick sentences directly from the data.

\textbf{Reddit.} Reddit forum data\footnote{https://www.reddit.com/} \footnote{http://discosumo.ruhosting.nl/wordpress/project-deliverables/} were prepared by Wubben et al.\cite{wubben2015facilitating}.  It contains 242,666 threads in 12,980 subreddits. The size of threads ranges from 5 sentences with a few words per line to over 43,000 sentences. In our study, we utilize threads with a length of at least 10 sentences, since any threads of smaller size no longer requires a summary. The training and test sets contain 66,589 and 17,869 threads respectively, while the development set contains 20,891 threads for hyperparameter-tuning. The reference summaries were prepared by using the number of votes as a factor to select sentences.  That is, all sentences are first ranked based on their final votes (No. of upvotes - No. of downvotes), then the ranked sentences are iteratively added into the output list until total words reach the compression ratio (25\% of original total words), and finally the selected sentences are ordered according to their chronological order.

\textbf{Newsroom.} Newsroom\footnote{https://lil.nlp.cornell.edu/newsroom} summarization dataset contains 1.3 million articles and summaries written by authors and editors in the newsrooms of 38 major publications \cite{grusky2018newsroom}. It is used for training and evaluating summarization systems. The dataset provides training, development, and test sets. Each set comprises summary objects, where each individual one includes information of article text, its corresponding summary, date, density bin, just to name a few. Density bin denotes summarization strategies of the reference summary, which involves extractive, abstractive, and a mix of both. In our study, we use only articles of which the reference summary was generated via extractive approach. Similar to Reddit dataset, we filter out any articles with number of sentences lower than 10. As a result, training and test sets contain a total of 288,671 and  31,426 articles respectively, while a development set contains 31,813 articles. 
\begin{table}[h]
\caption{Data statistics}
\centering
\begin{tabular}{c|c|c|c}
\hline

                 & \textbf{Trip advisor}                               &   \textbf{Reddit}                       & \textbf{Newsroom}           \\ \hline
Vocabulary                                                     & 26,422                               &   910,941                       & 894,212           \\
\#Threads                                                      & 700                               & 105,349                         & 351,910                           \\
\begin{tabular}[c]{@{}c@{}}Avg \#sentences\end{tabular}  & 59.41                               & 64.99                         & 35.57                           \\
\begin{tabular}[c]{@{}c@{}}Max \#sentences\end{tabular}      & 144                               & 43,486                         & 12,016                           \\
\begin{tabular}[c]{@{}c@{}}Avg \#words\end{tabular}      & 833.65                               & 783.85                         & 679.24                           \\
\begin{tabular}[c]{@{}c@{}}Max \#words\end{tabular}          & 1,559                               & 532,400                         & 178,463                           \\
\begin{tabular}[c]{@{}c@{}}Avg \#words per sentence\end{tabular}    & 14.03                               & 11.84                         & 16.23                           \\

\hline
\end{tabular}
\label{tab:stats}
\end{table}

\begin{table*}
\centering
\caption{Hyperparameter values evaluated in the proposed model.}
\begin{tabular}{l|l}
\hline
\multicolumn{1}{c|}{\textbf{Hyperparameter}} & \multicolumn{1}{c}{\textbf{Range}}                                                                                   \\ \hline
Number of Bi-LSTM hidden layer neurons       & \{25, 50, 100, 200\}                                                                                                 \\ 
Number of convolutional layers               & \{1, 2, 3, 4, 5\}                                                                                                             \\ 
Number of CNN filters                        & \{100, 200, 400, 600\}                                                                                               \\ 
CNN receptive field size                              & \{1, 2, 3, 4, 5, [2,3], [2,3,4], [2,3,4,5]\}                                                                                                          \\ 
Dropout rate                                 & \{0.0, 0.1, 0.3, 0.5\}                                                                                          \\ 
Learning rate                                & \{0.001, 0.01, 0.1\}                                                                                                 \\ 
Batch size                                   & \{16, 32, 64, 128\}                                                                                                  \\ 
Optimizer                                    & \begin{tabular}[c]{@{}l@{}}\{SGD\cite{ruder2016overview}, RMSprop\cite{hinton2012neural}, Adadelta\cite{DBLP:journals/corr/abs-1212-5701}, \\ Adagrad\cite{duchi2011adaptive}, Adam\cite{kingma2014adam}, Adamax\cite{kingma2014adam}\}\end{tabular} \\ 
\hline
\end{tabular}
\label{tab:hyperparam}
\end{table*}
\subsection{Training Set Creation}
In this study, every sentence requires a label to train the deep neural network; therefore, we create a training set where each sentence will be marked as True to indicate a part-of-summary unit, or False to indicate otherwise.
First of all, an empty set S = \{ \} is initialized per thread (or per news article). For each sentence that is not a member of the set, add the sentence to the set then measure ROUGE-1 score \cite{lin-2004-rouge} between the set and the gold summaries; thereafter, the sentence is removed from the set. Once all sentences had their ROUGE-1 score measured, the candidate sentences that increased the score the most are permanently added to the set. This process is repeated until one of the following conditions is achieved: 1) the total number of words in the selected sentences has hit the desired compression ratio of 20\%, or 2) the ROUGE score of summary cannot be improved any further. Finally, those sentences that are a member of the set are labeled True, while others are labeled False. We utilized the ROUGE 2.0 Java package\footnote{http://kavita-ganesan.com/content/rouge-2.0} to evaluate the ROUGE scores that presents the unigram overlap between the selected sentences and the gold summaries \cite{ganesan2015rouge}.

\subsection{Model Configuration}\label{model_config}
The optimum parameters for the proposed model were explored through experimentation. Six-fold cross-validation was used for both tuning and training process.  We performed a random search by sampling without replacement over 80\% of all possible configurations since the whole configuration space is too large. All hyperparameters are listed in Table \ref{tab:hyperparam}, some of which are based on the recommendation of \cite{zhang-wallace-2017-sensitivity}. We found the best configuration for the number of Bi-LSTM neurons at sentence encoder to be 200, and at thread encoder to be 100, respectively. For CNN hyperparameters, the best explored number of convolutional layers at sentence and thread levels is 2; and the best number of filters at both levels is 100, where each filter has the size and a stride length of 2. The best explored dropout rate is 0.3, with learning rate of 0.001 and a batch size of 16. Lastly, RMSprop optimizer has shown to best optimize binary cross-entropy loss function in our model. 

The training/validation/test split was set to 0.8/0.1/0.1 of all threads. We kept this split ratio fixed in all the experiments and all datasets. To prevent the model from overfitting, we applied early stopping during the training process. This was done by computing an error value of the model on a validation dataset for every epoch and terminating the training if the error value monotonically increased. After obtaining the best configuration, we retrained the model on the union of training and development sets and evaluated it on the test set. 
The training process continues until loss value converges or the maximum epoch of 500 is met. 

Regarding the pretrained vectors, we apply word2vec\footnote{https://code.google.com/archive/p/word2vec/}, FastText \cite{mikolov2018advances}\footnote{https://fasttext.cc/docs/en/english-vectors.html}, and ELMo\footnote{https://allennlp.org/elmo} as word-level embedding vectors. For BERT\footnote{https://github.com/google-research/bert}, however, we apply it as sentence-level embedding vectors since sentence vectors trained by BERT have shown to give better performances.

\subsection{Baselines}
We compare the proposed model against unsupervised and supervised methods. Detailed descriptions of each baseline are as follows.

\subsubsection{Unsupervised-learning Baselines} 
The following unsupervised-learning baselines are used for our comparative study:
\begin{itemize}
\item ILP ~\cite{berg-kirkpatrick-etal-2011-jointly}, a baseline Integer Linear Programming framework implemented by~\cite{boudin-etal-2015-concept}.

\item SumBasic ~\cite{Vanderwende:2007}, an approach that assumes words occurring frequently in a document cluster have a higher chance of being included in the summary.

\item KL-SUM ~\cite{haghighi-vanderwende-2009-exploring}, a method that adds sentences to the summary so long as it decreases the KL Divergence.

\item LSA ~\cite{steinberger2004using}, the latent semantic analysis technique to identify semantically important sentences. 

\item LEXRANK ~\cite{Erkan:2004}, a graph-based summarization approach based on eigenvector centrality.

\item MEAD ~\cite{Radev:2004}, a centroid-based summarization system that scores sentences based on sentence length, centroid, and position. 

\item Opinosis ~\cite{ganesan2010opinosis}, a graph based algorithm for generating abstractive summaries from large amounts of highly redundant text. 

\item TextRank ~\cite{barrios2016variations}, a graph-based extractive summarization algorithm which computes similarity among sentences.

\end{itemize}

\subsubsection{Supervised-learning Baselines}
We also include traditional supervised-learning methods namely Support Vector Machine (SVM) and LIBLINEAR \cite{Fan:2008}\footnote{Set option `-c 10 -w1 5' for SVM and `-c 0.1 -w1 5' for LogReg} in our study. Both of which employ the following features: 1) cosine similarity of current sentence to thread centroid, 2) relative sentence position within the thread, 3) the number of words in the sentence excluding stopwords, and 4) max/avg/total TF-IDF scores of the consisting words. The features were designed such that they carry similar information as our proposed model.

\subsubsection{Deep Learning Baseline}
Neural network methods, including LSTM and CNN, have been used as a deep learning baseline in our study. 
For LSTM, we implemented a neural network containing a single layer of LSTM to classify sentences in each input thread/news article. 
For CNN, the sentence classification model proposed by \cite{kim-2014-convolutional} is applied. The input layer was initialized with pre-trained static word embeddings. The network uses features extracted from the convolutional layer for the classification. 

In addition, we implemented a variant of HAN, namely hierarchical convolutional neural network (HCNN) which employs CNN rather than LSTM. This allows us to examine the effectiveness of each individual network.

\subsection{Evaluation Methods}
We report ROUGE-1, ROUGE-2, and ROUGE-L scores along with sentence-level scores for the evaluation. The quantitative values for each method are computed as precision, recall, and F1 measure. Note that we will also refer to ROUGE metrics as R-1, R-2, and R-L for short. 

ROUGE-1 and ROUGE-2 are metrics commonly used in the DUC and TAC competitions to evaluate the quality of system summary \cite{Dang:2008}. Their precision scores are computed as the number of n-grams the system summary has \textit{in common} with its corresponding human reference summaries divided by total n-grams in the system summary where R-1 and R-2 set n=1 and n=2, respectively. R-1 and R-2 recall scores are calculated the same way except that the number of overlapping n-grams are divided by the total n-grams in the human reference summary. Finally, the F1 score for R-1 and R-2 is the harmonic mean of precision and recall. We use R-1 and R-2 as a means to assess informativeness. ROUGE-L measures the longest common subsequence of words between the sentences in the system summary and the reference summary. The higher the R-L, the more likely that the output summary has n-grams in the same order as the reference summary. In other words, R-L indicates how well the output summary preserves the semantic meaning of the reference summary. 

Sentence-level scores report the classification performance of the model.
Each sentence is labelled true or false to signify if it is part of the summary.
When the label is true in both reference set (actual class) and system set (predicted class), this case is considered true positive.
If labelled true in the reference set yet false in the system set, this case is considered false negative. 
On the other hand, if labelled true in the system set yet false in the reference set, this case is considered false positive. 
Finally, a sentence labelled false in both the system set and the reference set is considered true negative. Table \ref{tab:conf_matrix} presents the confusion matrix.
\begin{table}[h]
\begin{center}
\caption{Confusion Matrix.}
\begin{tabular}{cc|cc}
\hline
                                                                                                       &                & \multicolumn{2}{c}{\textbf{Predicted Class}}                                                                                                         \\  
                                                                                                       &                & \multicolumn{1}{c|}{\textbf{True}}                                                  & \textbf{False}                                                 \\ \hline
\multicolumn{1}{c}{\multirow{2}{*}{\textbf{\begin{tabular}[c]{@{}c@{}}Actual \\ Class\end{tabular}}}} & \textbf{True}  & \multicolumn{1}{c|}{\begin{tabular}[c]{@{}c@{}}True Positives\\ (TP)\end{tabular}}  & \begin{tabular}[c]{@{}c@{}}False Negatives\\ (FN)\end{tabular} \\ \cline{2-4} 
\multicolumn{1}{c}{}                                                                                  & \textbf{False} & \multicolumn{1}{c|}{\begin{tabular}[c]{@{}c@{}}False Positives\\ (FP)\end{tabular}} & \begin{tabular}[c]{@{}c@{}}True Negatives\\ (TN)\end{tabular}  \\ 
\hline
\end{tabular}
\label{tab:conf_matrix}
\end{center}
\end{table}

Sentence-level precision is the number of true positives divided by the sum of true positives and false positives. Sentence-level recall is the number of true positives divided by the sum of true positives and false negatives. Lastly, sentence-level F1 is the harmonic mean of recall and precision. 

\section{Performance Evaluation Results and Discussions}\label{sec:results}
Our proposed network is compared against a set of unsupervised and supervised approaches, along with variants of hierarchical methods. In this section, we first discuss the performance of different methods which involve traditional machine learning baselines, non-hierarchical and hierarchical deep learning methods. Then, we explain comparisons, observations, and provide our detailed analysis. After that, extensive ablation studies are presented.

\subsection{Comparison with Traditional Machine Learning Baselines}
Among the unsupervised-learning baselines in Table \ref{tab:main_sent_level_results}, the sentence classification results from MEAD demonstrate good performance. MEAD has also been shown to perform well in previous studies such as \cite{luo-etal-2016-automatic}. In this study, MEAD and LexRank are centroid-based, meaning that sentences that contain more words from the cluster centroid are considered to be holding key information, thereby increasing the likelihood of being included in the final summary. A similar pattern in results appears in KL-Sum and LSA. Nonetheless, in terms of ROUGE evaluation as shown in Table \ref{tab:main_summ_results}, they were all outperformed by hierarchical-based deep learning approaches. Opinosis has poor performance since it relies heavily on the redundancy of the data to generate a graph with meaningful paths. To this end, the hierarchical approaches appear to achieve better performance without the need for sophisticated constraint optimization such as in ILP.

Regarding the supervised-learning baselines, according to Table \ref{tab:main_summ_results}, a pattern of high precision and low recall can generally be observed for both SVM and LogReg. The R-1 results reflect that among the sentences classified as True, there are several unigrams overlapping with the reference summaries. However, when evaluating with higher n-grams, the results show that only a few matches exist between the system and the references. Considering the sentence-level scores of the Trip advisor dataset as an example, it can be seen that LogReg has failed to extract representative sentences as evidenced from the 14.50\% precision and 12.10\% recall, which are the lowest. 

The comparison of traditional models against hierarchical-based models has shown that the hierarchical models have better potential in classifying and selecting salient sentences to form a summary. Furthermore, both traditional machine learning baselines possess one disadvantage which is their reliance on a set of features from the feature engineering process. These handcrafted features might not be able to capture all the traits necessary for the models to differentiate between classes.

\subsection{Comparison to Non-hierarchical Deep Learning Methods}
In general, LSTM outperforms CNN in terms of sentence classification as well as ROUGE evaluation. Particularly for the sentence classification task, LSTM has shown to achieve high precision scores across all datasets. This indicates the importance of the learning of sequential information towards obtaining an effective representation. CNN, although proven to be efficient in previous studies, the results have evidenced that omitting sequential information essentially results in an inferior performance as shown in Table \ref{tab:main_sent_level_results}.

In terms of ROUGE evaluation, according to Table \ref{tab:main_summ_results}, R-1 and R-2 of both LSTM and CNN baselines are quite competitive compared to the hierarchical-based methods. However, with respect to R-L scores, hierarchical-based models generally have better performance by a significant margin. We observe that, hierarchical models have an advantage over the non-hierarchical deep learning methods in that they also explore hierarchical structure on top of sequential information learned via LSTM or feature extraction learned via CNN.

\begin{figure*}[t!]
      \centering
  \begin{subfigure}{6cm}
    \centering\includegraphics[width=\textwidth]{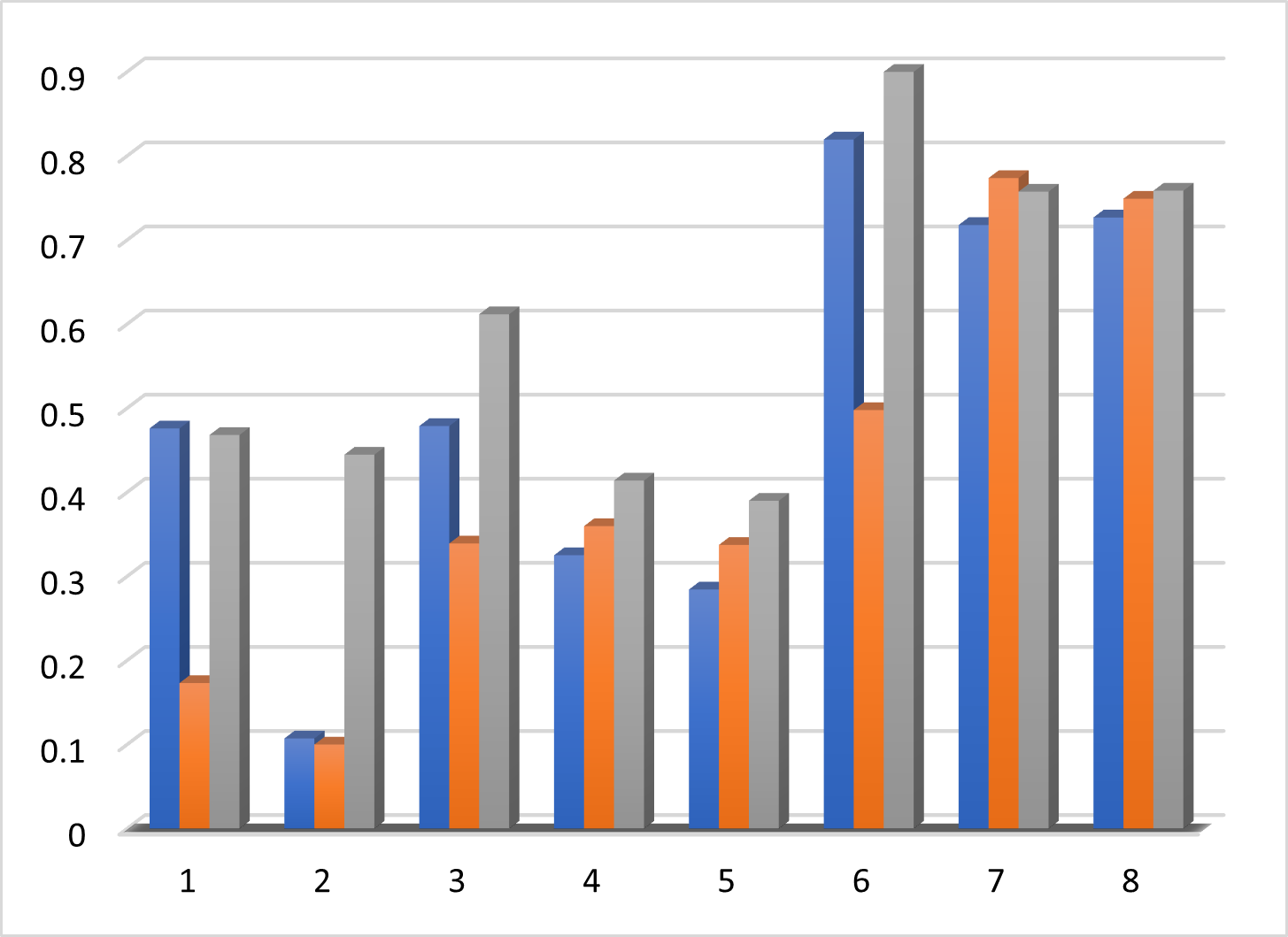}
    \caption{Trip advisor}
  \end{subfigure}
  \begin{subfigure}{6cm}
    \centering\includegraphics[width=\textwidth]{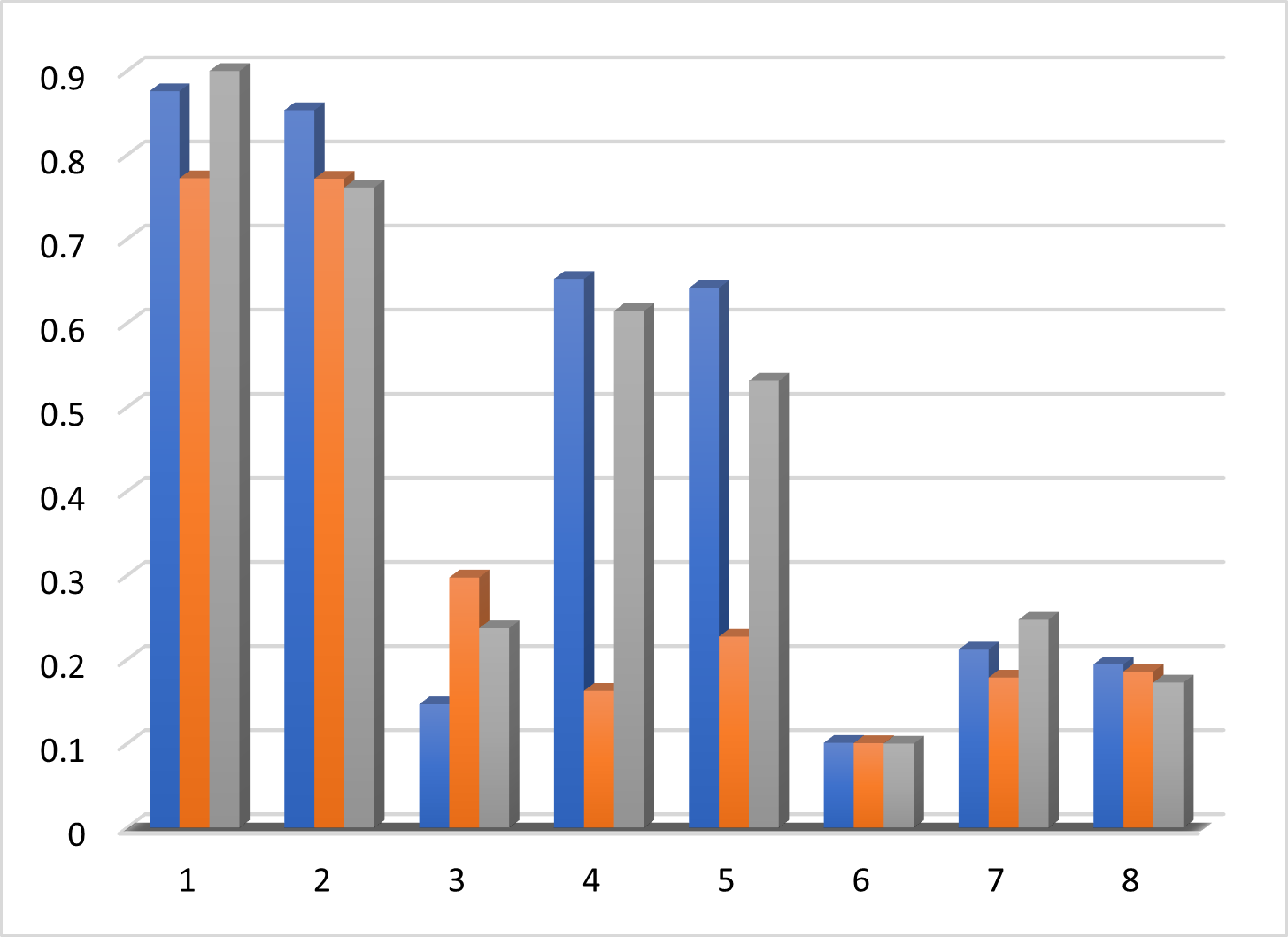}
    \caption{Reddit}
  \end{subfigure}
  \begin{subfigure}{6cm}
    \centering\includegraphics[width=\textwidth]{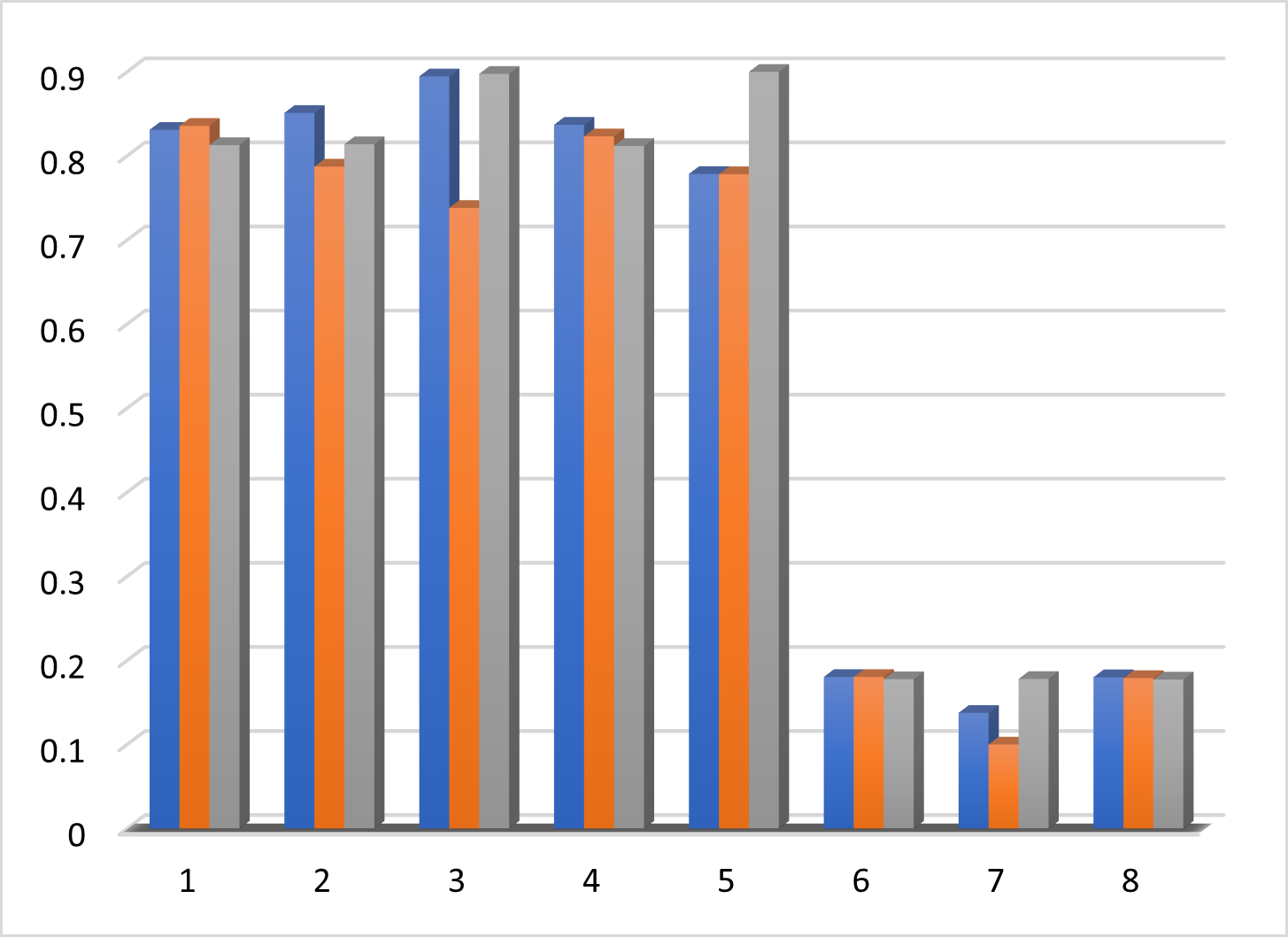}
    \caption{Newsroom}
  \end{subfigure}
  \caption{Comparison of F1-scores among Hierarchical Methods based on \textit{sentence-level scores}. \textbf{x-axis} denotes types of embeddings; 1=w2v, 2=FastText, 3=ELMo, 4=ELMo+w2v, 5=ELMo+FastText, 6=BERT, 7=BERT+w2v, 8=BERT+FastText. \textbf{y-axis} denotes F1-scores normalized between [0,1]. The bar color \textcolor{blue}{blue} presents HAN, \textcolor{orange}{orange} presents HCNN, and \textcolor{gray}{gray} presents the proposed model.}
    \label{fig:sent_level_bar}
\end{figure*}

\subsection{Comparison with Hierarchical Attention-based Deep Networks}\label{sec:compare_hier}
Of all the hierarchical-based models, we compare the proposed model against the state-of-the-art HAN model to examine whether the proposed architecture contributes to performance gain/loss. We hypothesize that both LSTM and CNN encourage the leverage of long-term and short-term dependencies, which are keys to learning and generating effective representation for the summarizer. We also make a comparison with the hierarchical convolutional neural network (HCNN) to observe the effect of excluding long-term dependencies captured by LSTM.

According to Table \ref{tab:main_sent_level_results}, the sentence-level score shows that on average the performance of our proposed network is comparative to other hierarchical methods regardless of the choice of embedding. HCNN is generally the most inferior among the three hierarchical models. This demonstrates that LSTM layers play a key role in capturing sequential information which is essential for the system to understand input documents. 
Without LSTM layers, the system only obtains high-level representation through CNN which is insufficient to generate an effective representation.
Using both LSTM and CNN has shown a promising avenue to improve the summarization task. 

It is important to note that, when the contextual representation is employed, especially for Reddit and Newsroom datasets, their results have shown high precision yet low recall. This indicates that few sentences are predicted as a part-of-summary sentence; however, most of its predicted labels are correct. Figure \ref{fig:sent_level_bar} illustrates a comparison among hierarchical methods with respect to sentence-level scores across all datasets.

\begin{figure*}
\centering
  \begin{subfigure}{6cm}
    \centering\includegraphics[width=\textwidth]{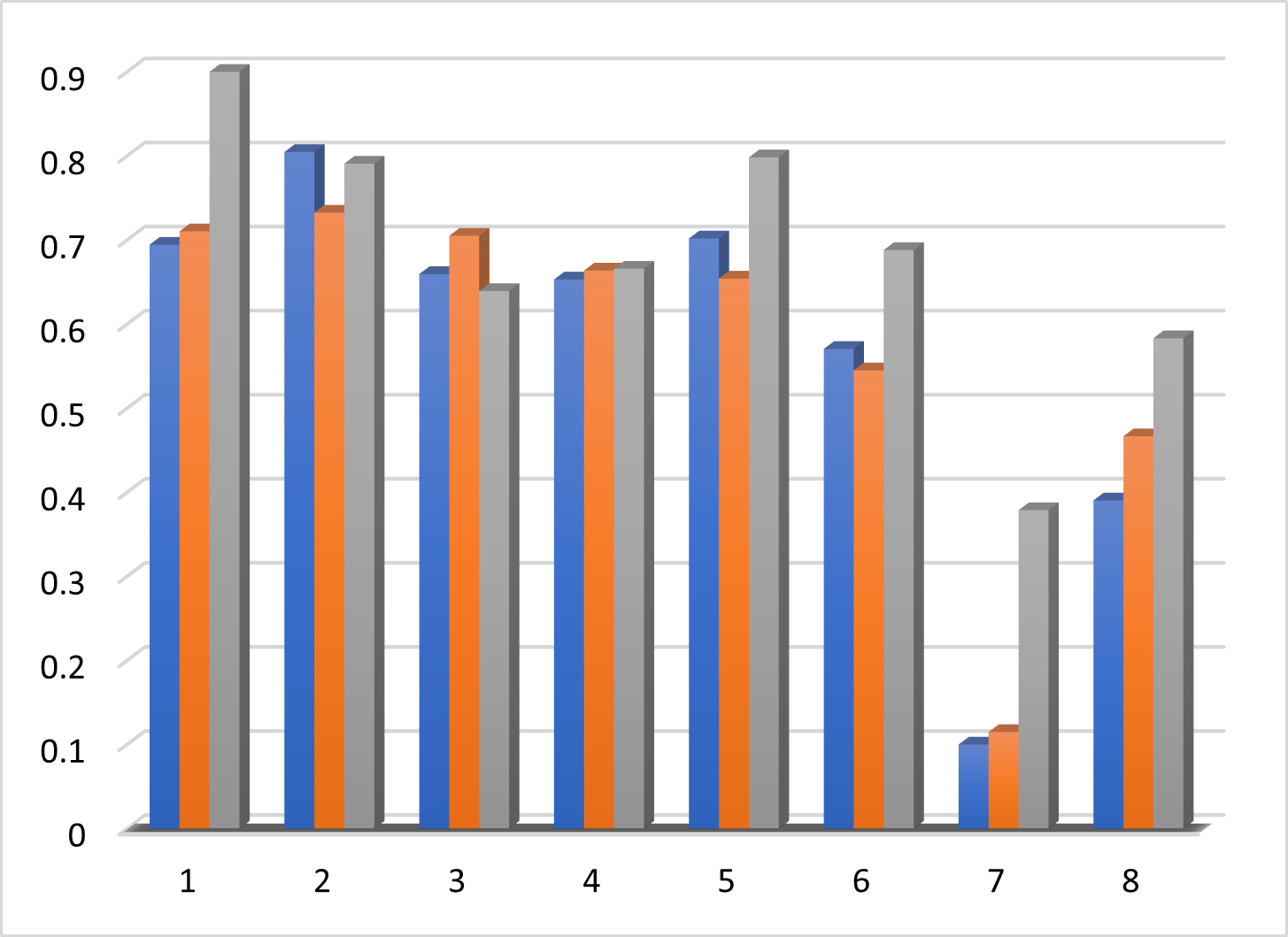}
    \caption{R-1, \textit{Trip advisor}}
  \end{subfigure}
  \begin{subfigure}{6cm}
    \centering\includegraphics[width=\textwidth]{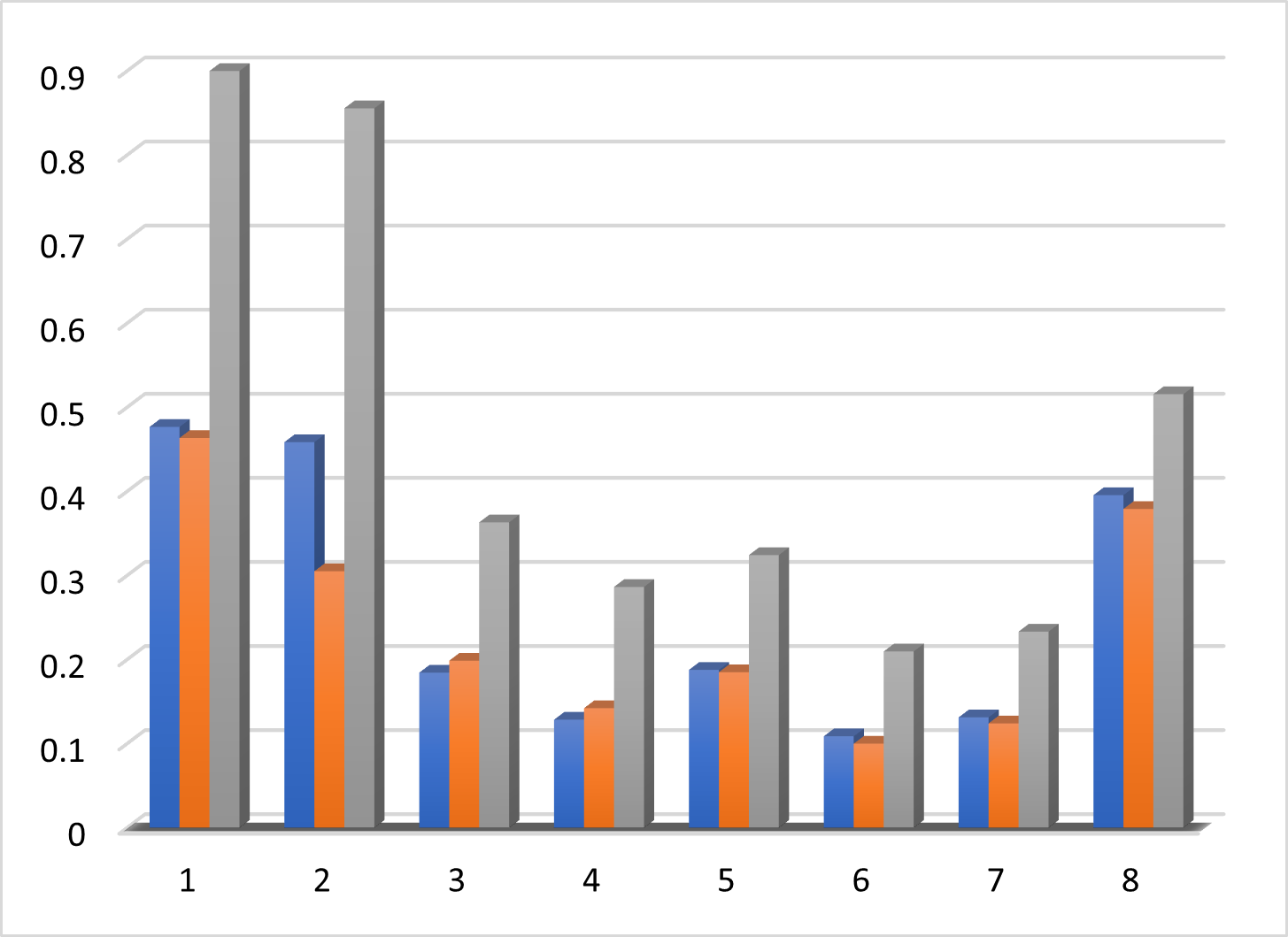}
    \caption{R-1, \textit{Reddit}}
  \end{subfigure}
  \begin{subfigure}{6cm}
    \centering\includegraphics[width=\textwidth]{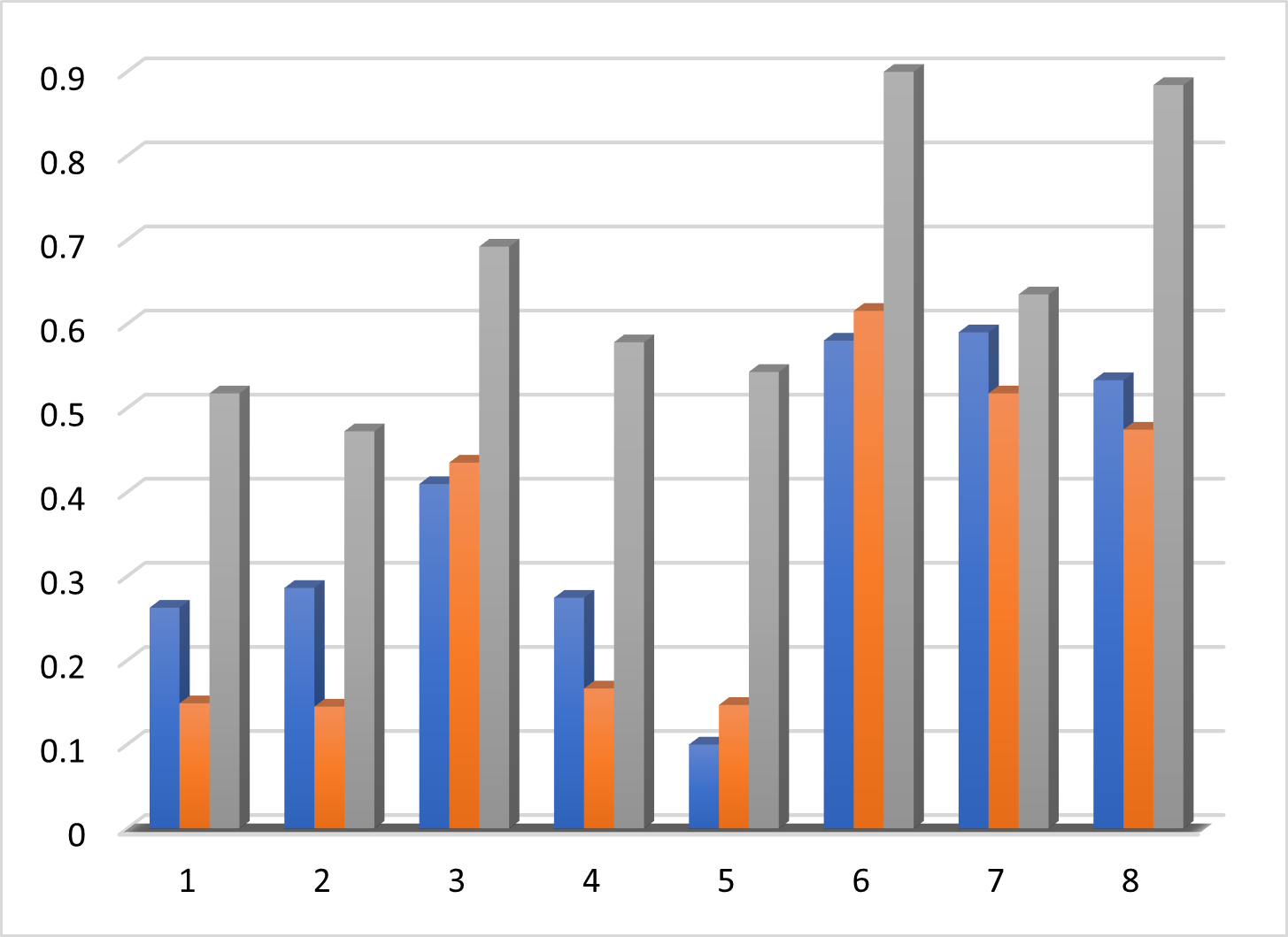}
    \caption{R-1, \textit{Newsroom}}
  \end{subfigure}
 \hspace{0.3\textwidth}
 
  \begin{subfigure}{6cm}
    \centering\includegraphics[width=\textwidth]{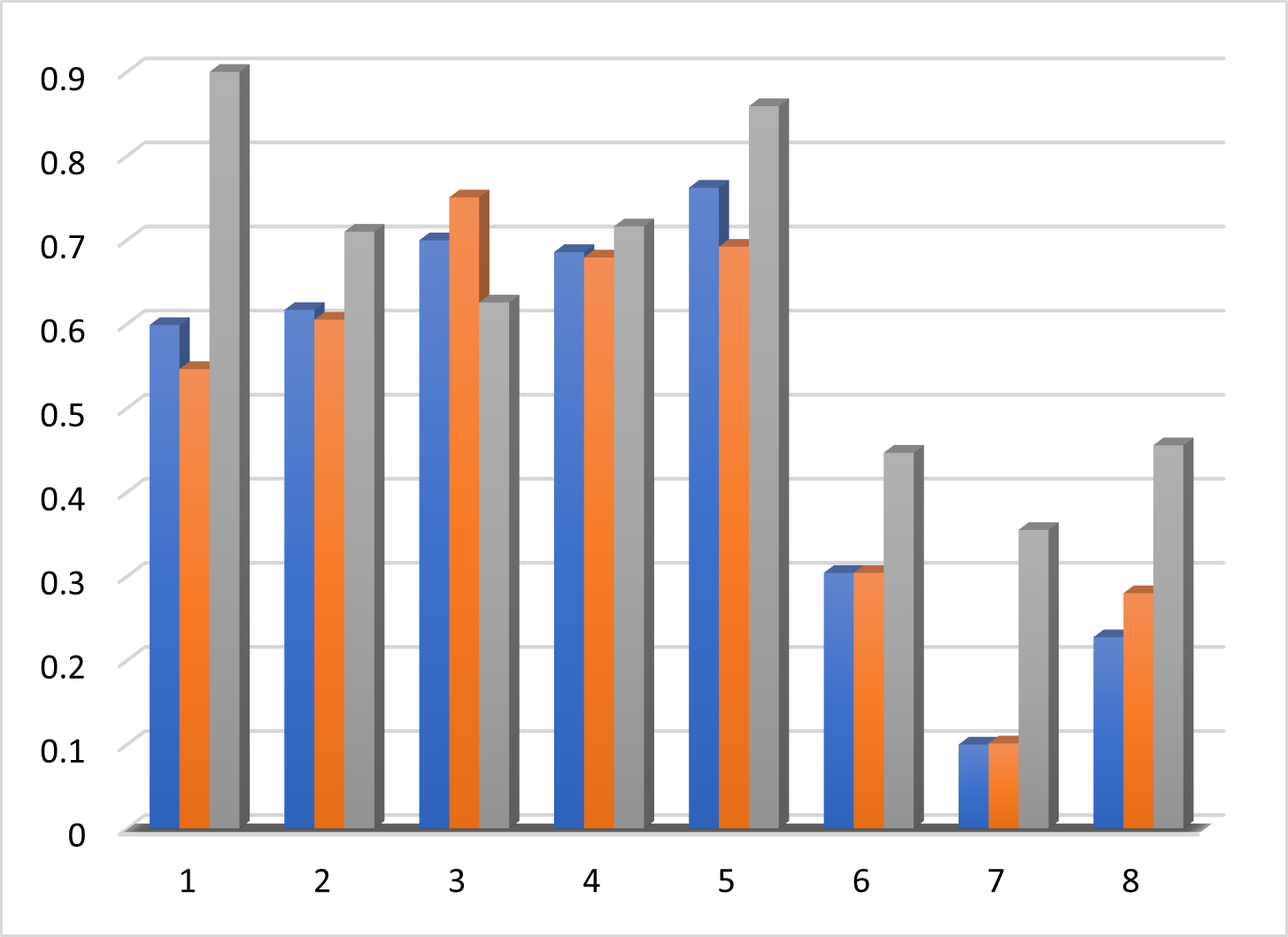}
    \caption{R-2, \textit{Trip advisor}}
  \end{subfigure}
  \begin{subfigure}{6cm}
    \centering\includegraphics[width=\textwidth]{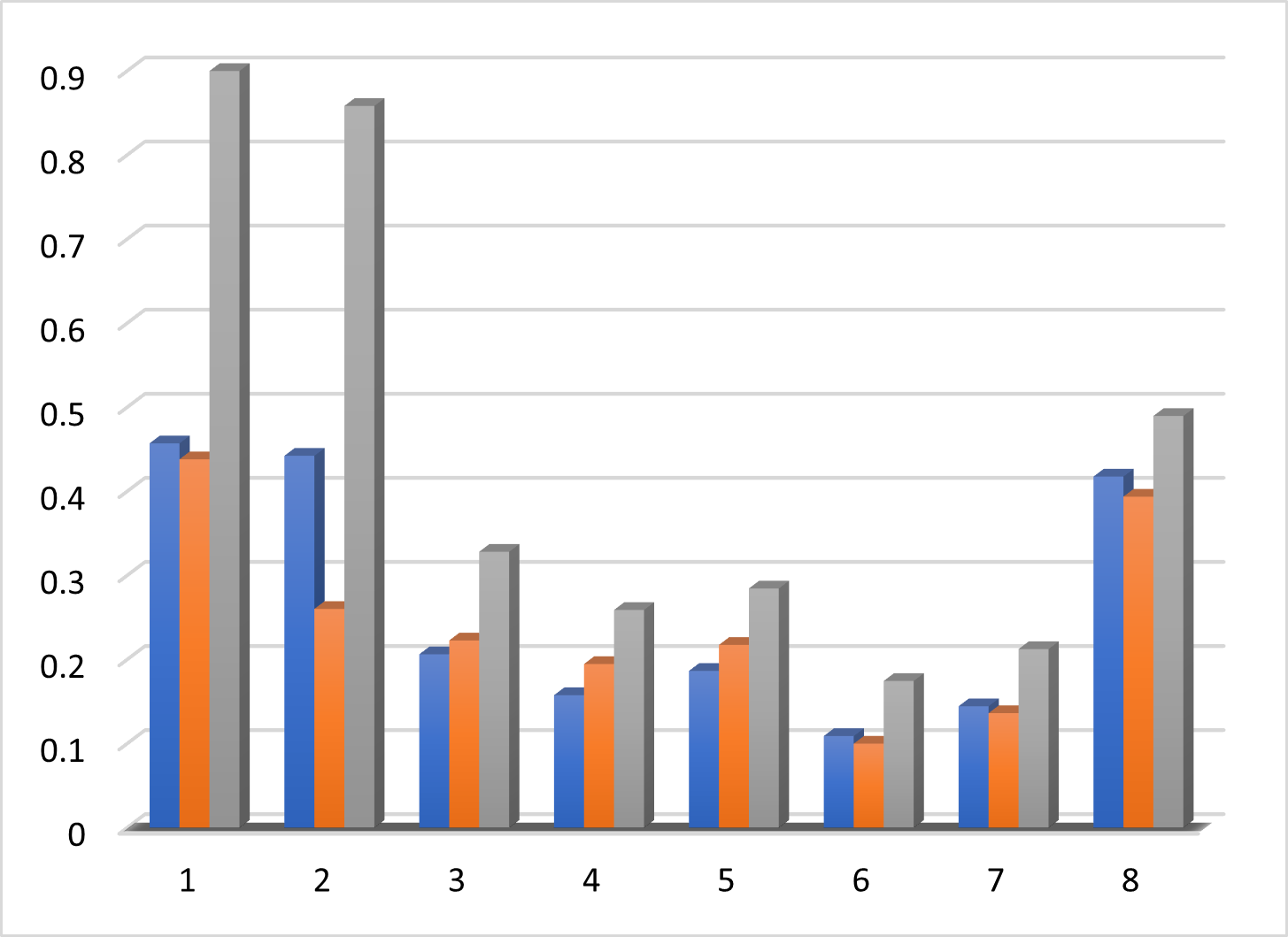}
    \caption{R-2, \textit{Reddit}}
  \end{subfigure}
  \begin{subfigure}{6cm}
    \centering\includegraphics[width=\textwidth]{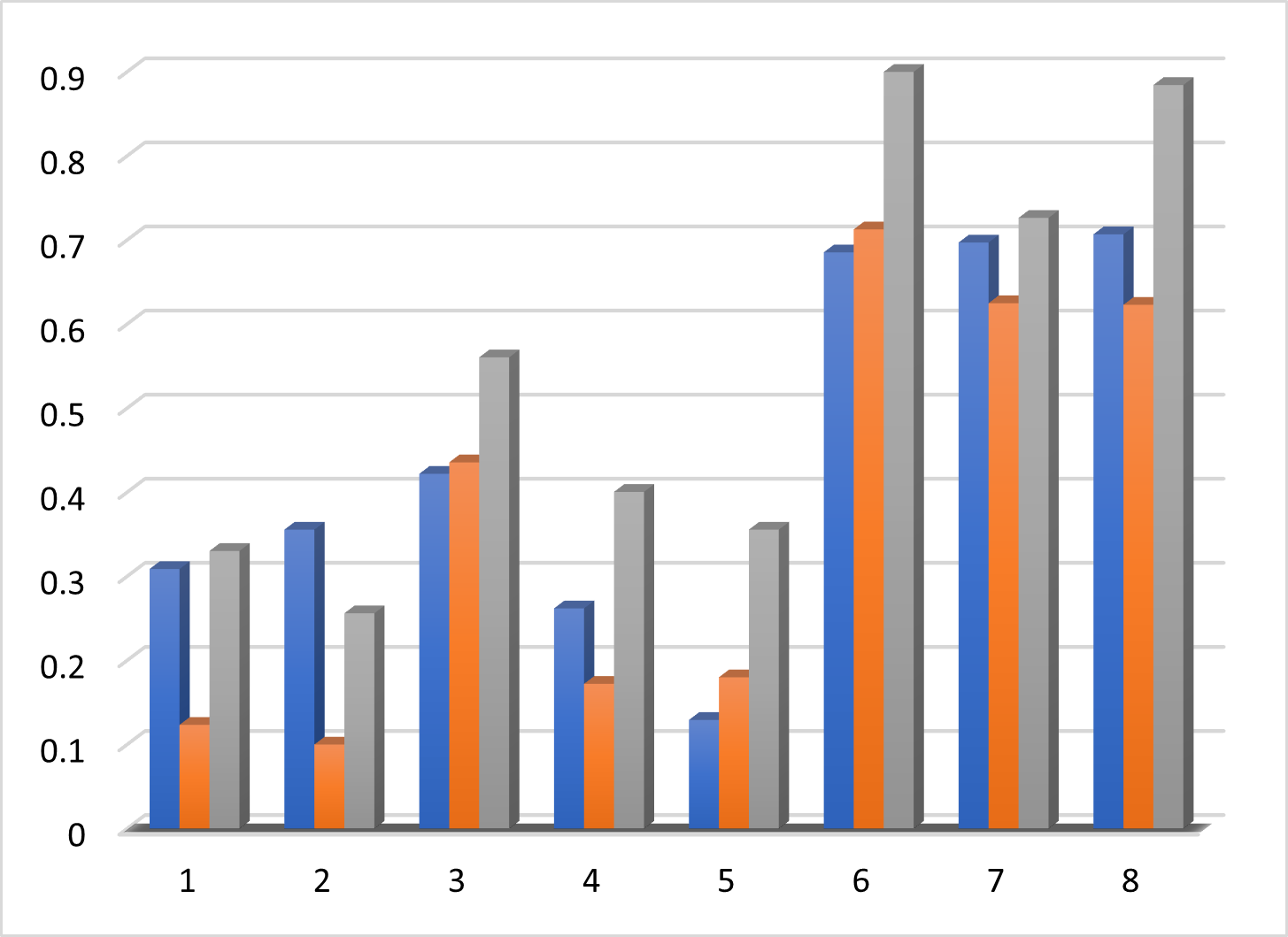}
    \caption{R-2, \textit{Newsroom}}
  \end{subfigure}
  
  \begin{subfigure}{6cm}
    \centering\includegraphics[width=\textwidth]{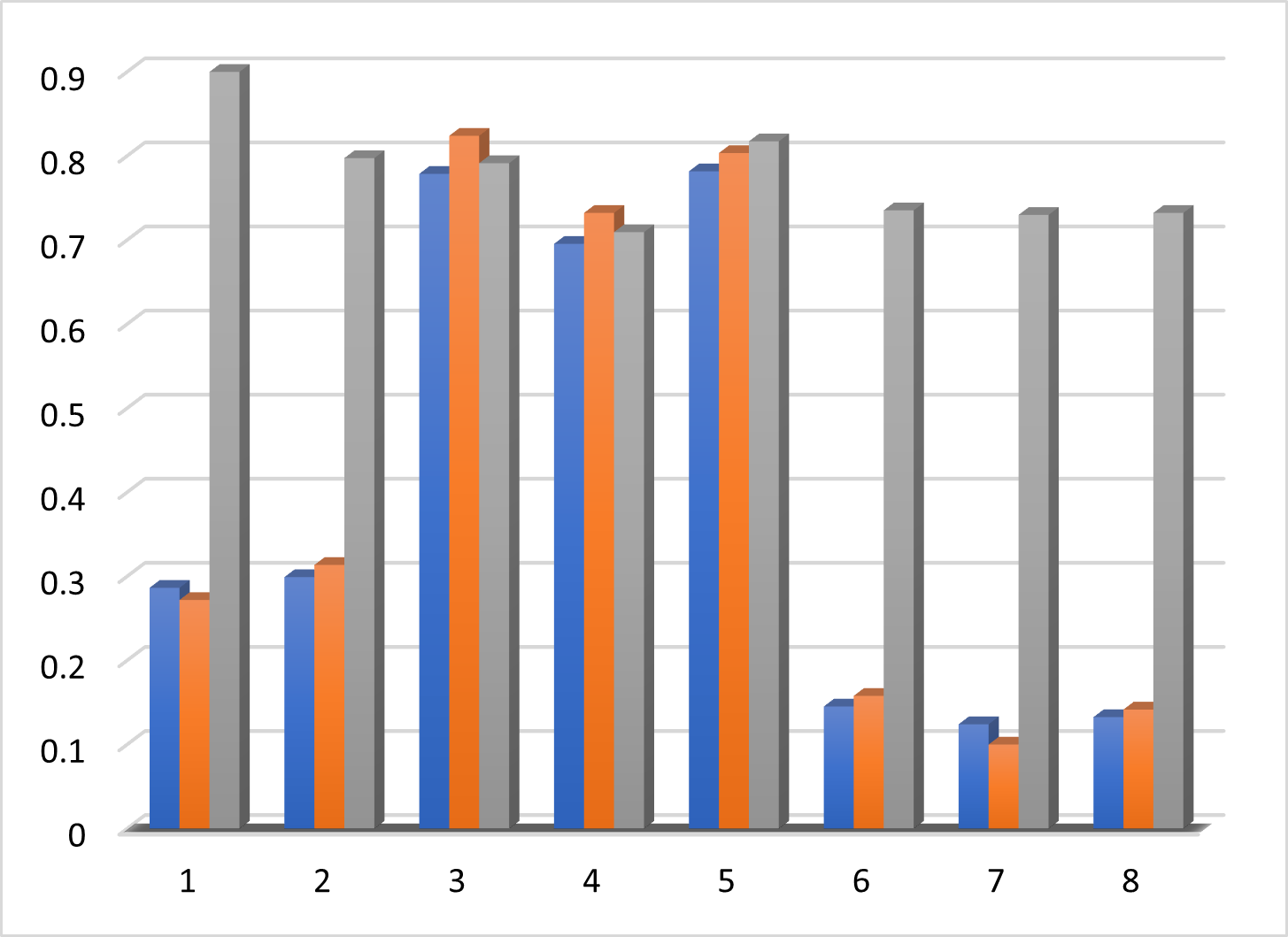}
    \caption{R-L, \textit{Trip advisor}}
  \end{subfigure}
  \begin{subfigure}{6cm}
    \centering\includegraphics[width=\textwidth]{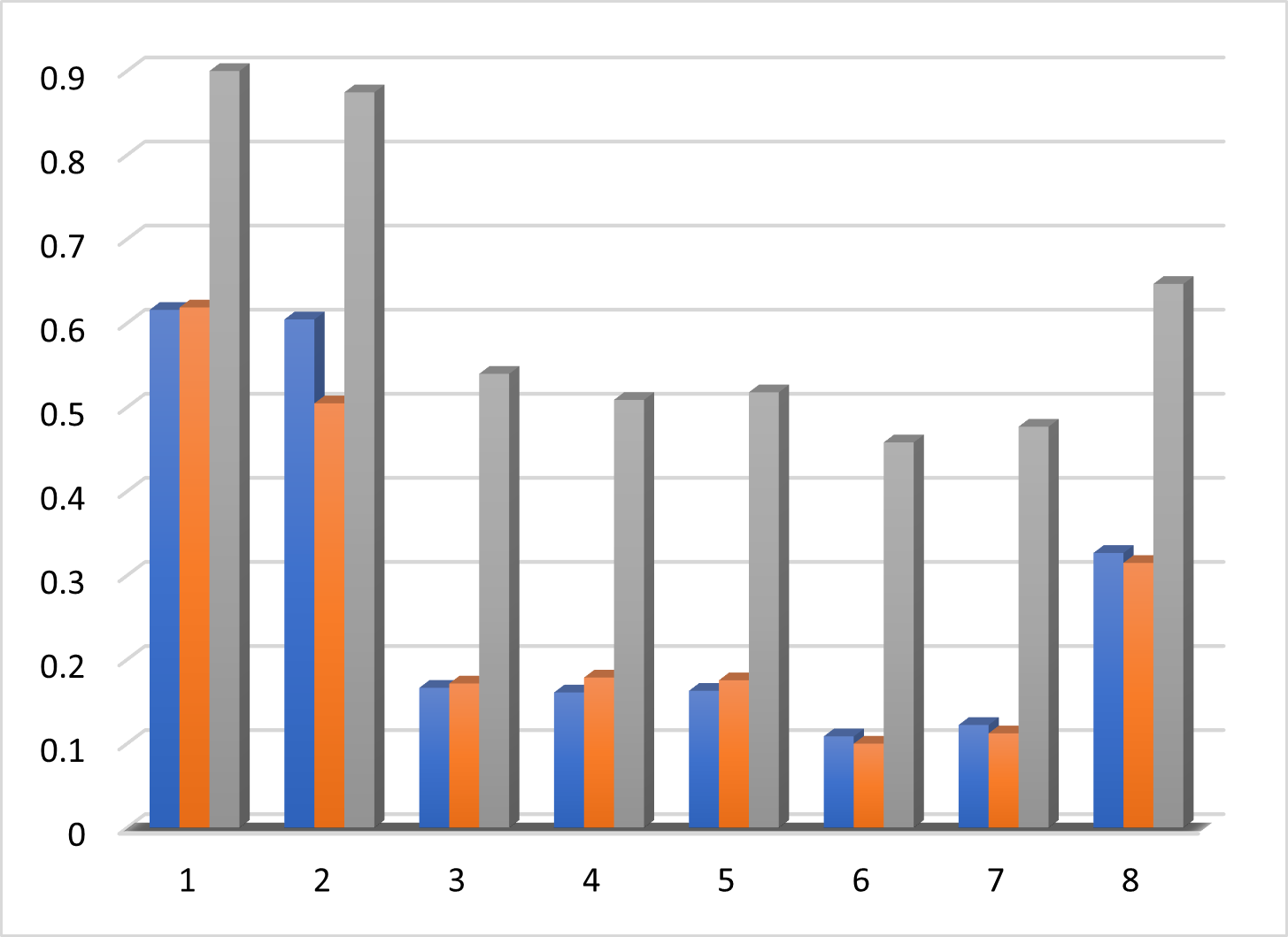}
    \caption{R-L, \textit{Reddit}}
  \end{subfigure}
  \begin{subfigure}{6cm}
    \centering\includegraphics[width=\textwidth]{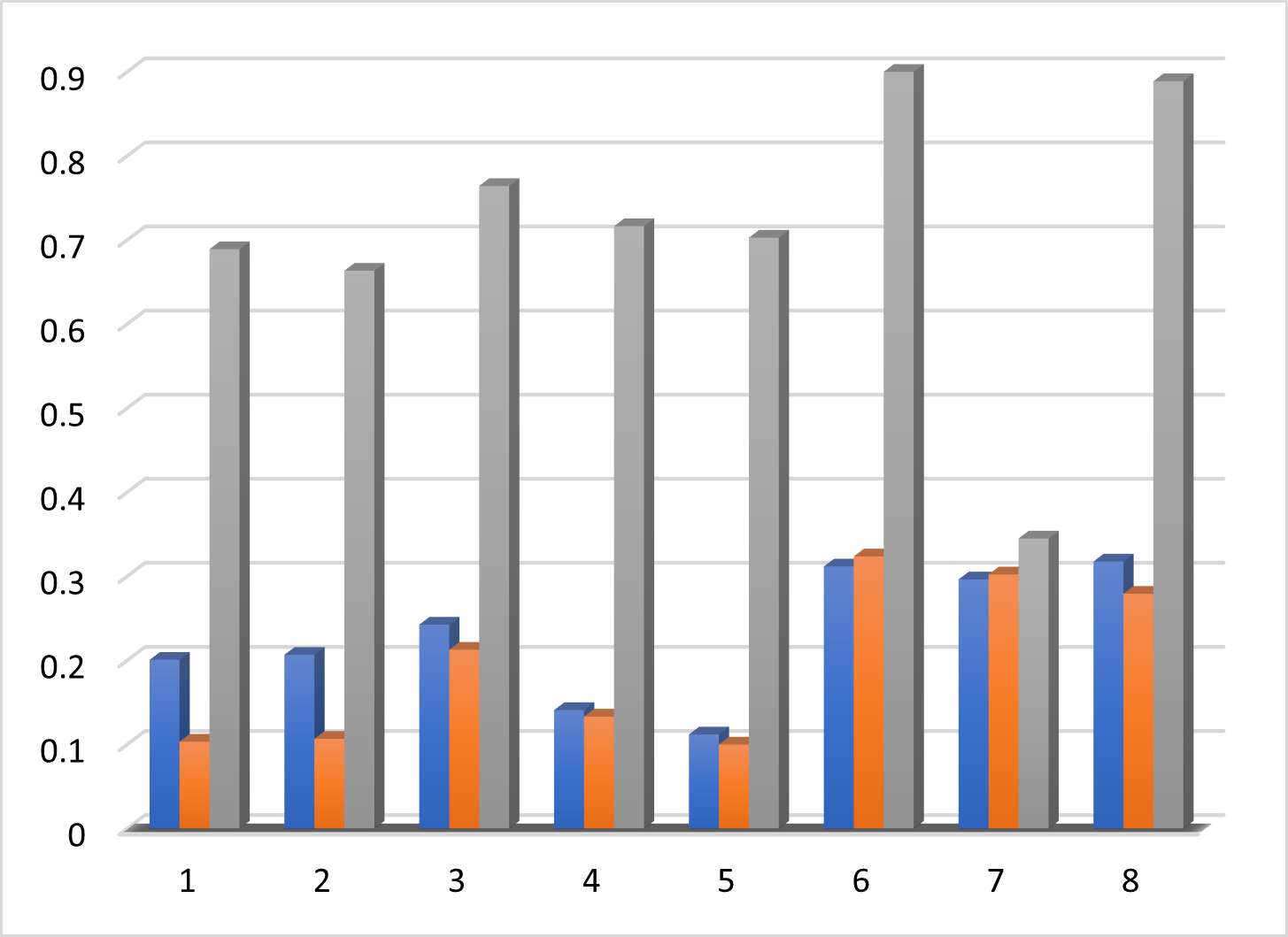}
    \caption{R-L, \textit{Newsroom}}
  \end{subfigure}
 
 \caption{Comparison of F1-scores among Hierarchical Methods based on \textit{ROUGE scores}. \textbf{x-axis} denotes types of embeddings; 1=w2v, 2=FastText, 3=ELMo, 4=ELMo+w2v, 5=ELMo+FastText, 6=BERT, 7=BERT+w2v, 8=BERT+FastText. \textbf{y-axis} denotes F1 scores normalized between [0,1]. The bar color \textcolor{blue}{blue} presents HAN, \textcolor{orange}{orange} presents HCNN, and \textcolor{gray}{gray} presents our proposed model. Fig. (4)(a)-(c) are R-1 scores, Fig. (4)(d)-(f) are R-2 scores, and Fig. (4)(g)-(i) are R-L scores.}
 \label{fig:bar_graphs}
\end{figure*}

\begin{figure*}
\centering
\includegraphics[width=0.8\textwidth]{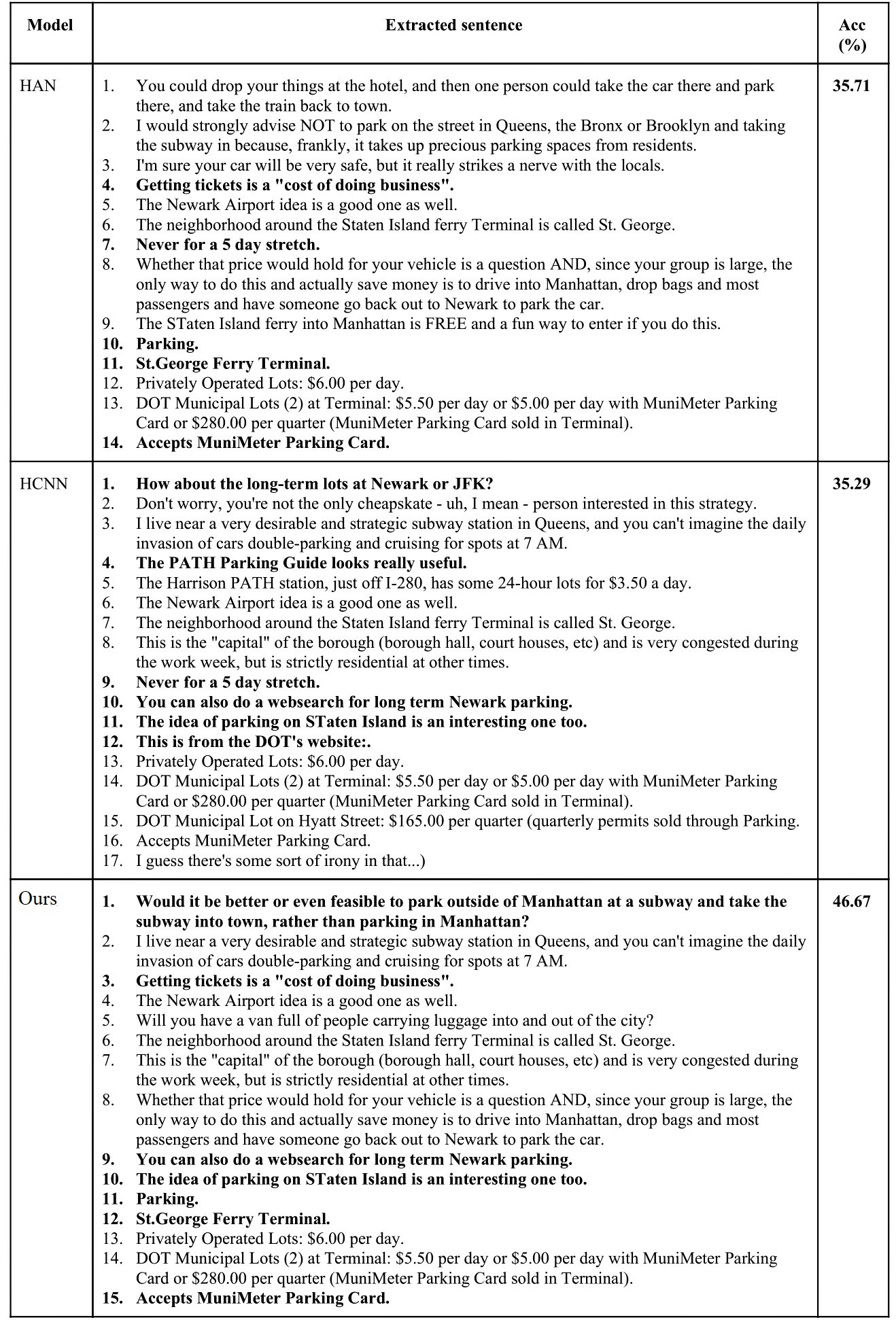} 
\caption{Example of output summaries generated by each hierarchical model.
Presented in \textbf{Bold} are correctly labelled sentences.
Accuracy value (\%) is computed as a ratio of number of correctly labelled sentences out of total sentences selected by the model.}
\label{fig:extracted_sents}
\end{figure*}

With respect to ROUGE evaluation, Table \ref{tab:main_summ_results} shows that ROUGE scores for the hierarchical models are promising. Among the hierarchical models, our proposed method outperforms others in all datasets, as displayed in Figure \ref{fig:bar_graphs} (a) - (i). We present example summaries generated by the hierarchical models in Figure \ref{fig:extracted_sents}. The results indicate that for our proposed model, among all its true-labeled sentences, 46.67\% were labeled correctly which is higher than the rest of the hierarchical models. 

\begin{figure*}
\centering
  \begin{subfigure}{8cm}
    \centering\includegraphics[width=\textwidth]{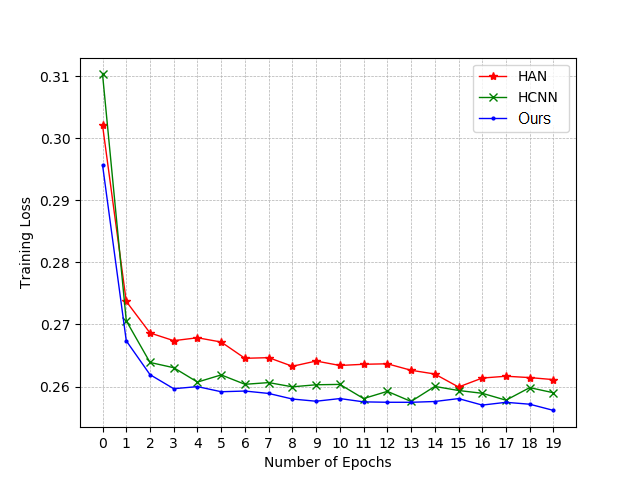}
    \caption{Fold 1}
  \end{subfigure}
  \begin{subfigure}{8cm}
    \centering\includegraphics[width=\textwidth]{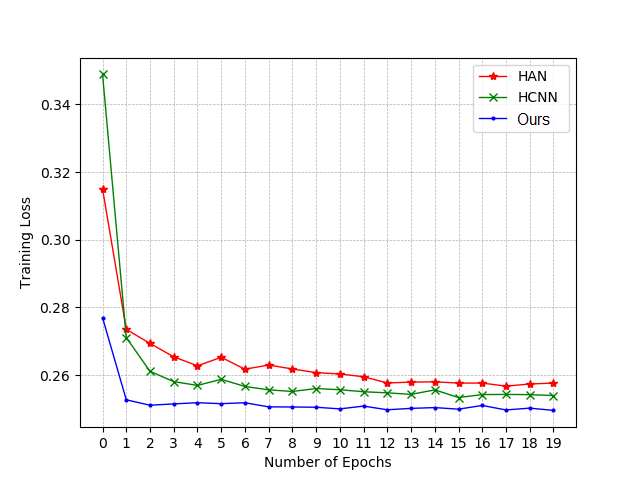}
    \caption{Fold 2}
  \end{subfigure}
 
  \begin{subfigure}{8cm}
    \centering\includegraphics[width=\textwidth]{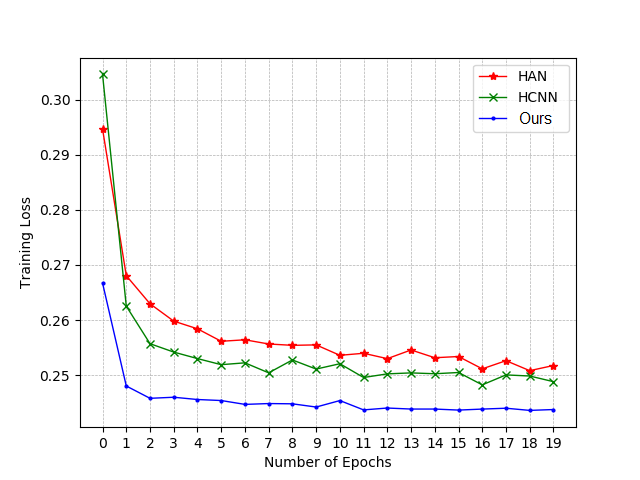}
    \caption{Fold 3}
  \end{subfigure}
  \begin{subfigure}{8cm}
    \centering\includegraphics[width=\textwidth]{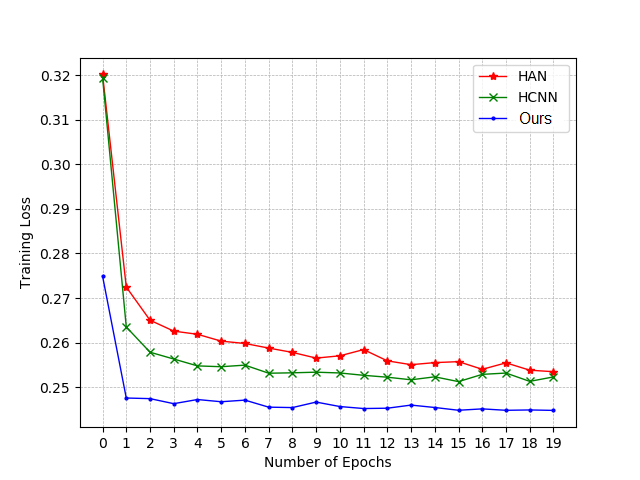}
    \caption{Fold 4}
  \end{subfigure}
  
    \begin{subfigure}{8cm}
    \centering\includegraphics[width=\textwidth]{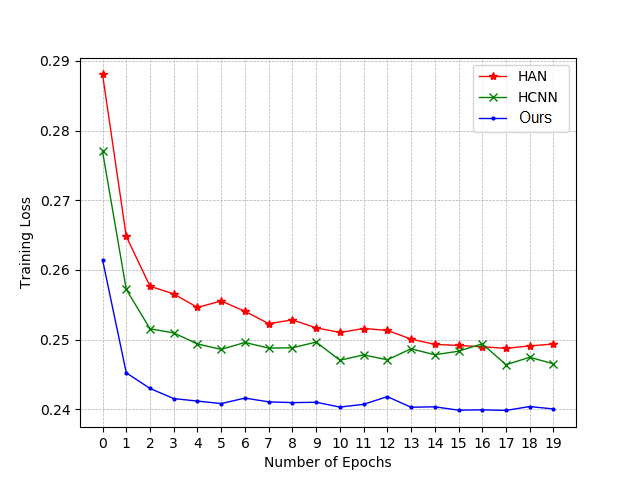}
    \caption{Fold 5}
  \end{subfigure}
  \begin{subfigure}{8cm}
    \centering\includegraphics[width=\textwidth]{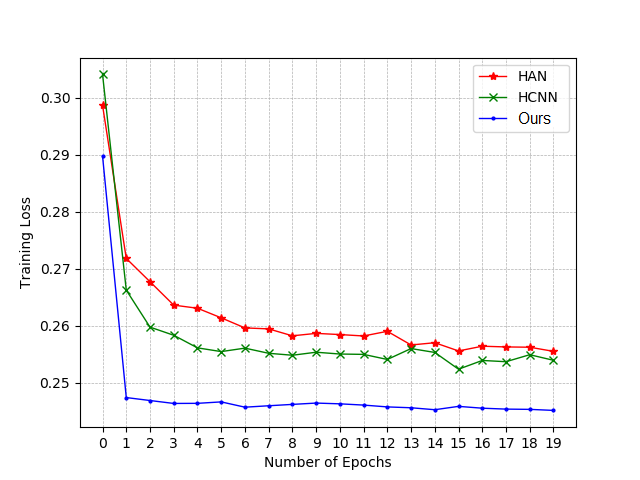}
    \caption{Fold 6}
  \end{subfigure}
 \caption{Plots showing the convergence of training loss per fold. The results from first 20 epochs are displayed since all models have shown to plateau from this point forward.}
 \label{fig:loss_converge_per_fold}
\end{figure*}

We also observed the behavior of each hierarchical model in terms of loss that is minimized. Figure \ref{fig:loss_converge_per_fold} (a) - (f) illustrate the training loss of each hierarchical model per fold. We note that for every fold of every model, the objective loss continuously decreases and begins to converge very early on. The average losses across all epochs of HAN, HCNN, and our proposed model are approximately 0.2555, 0.2564, and 0.2466, respectively. More fluctuations also appear in the HCNN curve. Our model converges faster due to its larger model complexity.

\begin{table*}
\setlength{\tabcolsep}{10pt}
\caption{Sentence-level classification results from all models. 
Precision (P), Recall (R), and F1 scores (F) are reported in percentage.
Variance ('\textpm') of F1 scores across all data are also presented.}
\begin{adjustbox}{width=\textwidth}
\begin{small}
\begin{tabular}{c||c|ccc|ccc|ccc}
\hline
                                    & \multicolumn{1}{c||}{}                   & \multicolumn{3}{c||}{\textbf{Trip advisor}}                & \multicolumn{3}{c||}{\textbf{Reddit}}                      & \multicolumn{3}{c}{\textbf{Newsroom}} \\ 
                                    \cline{3-11}
\textbf{Embedding}                  			& \multicolumn{1}{c||}{\textbf{Method}}    & \textbf{P} 			& \textbf{R} 		& \multicolumn{1}{c||}{\textbf{F}} 			& \textbf{P} 		& \textbf{R} 		& \multicolumn{1}{c||}{\textbf{F}} 			& \textbf{P}  		& \textbf{R}  			& \textbf{F} \\ \hline
\multicolumn{11}{c}{{\cellcolor[gray]{.9}}Baselines} \\ \hline
\multirow{8}{*}{            }       & \multicolumn{1}{c||}{ILP}       & 22.60      & 13.60      & \multicolumn{1}{c||}{15.60\textpm0.40}      & 22.86      & 23.40      & \multicolumn{1}{c||}{23.12\textpm0.10}      & 16.05       & 20.15       & 17.87\textpm0.10      \\
                                    & \multicolumn{1}{c||}{Sum-Basic} & 22.90      & 14.70      & \multicolumn{1}{c||}{16.70\textpm0.50}      & 22.31      & 17.11      & \multicolumn{1}{c||}{19.37\textpm0.20}      & 16.57       & 18.52       & 17.49\textpm0.10      \\
                                    & \multicolumn{1}{c||}{KL-Sum}    & 21.10      & 15.20      & \multicolumn{1}{c||}{16.30\textpm0.50}      & 23.58      & 17.91      & \multicolumn{1}{c||}{20.36\textpm0.10}      & 17.25       & 20.74       & 18.83\textpm0.20      \\
									& \multicolumn{1}{c||}{LSA}   	& 21.05      & 15.02      & \multicolumn{1}{c||}{17.53\textpm0.50}      & 23.59      & 17.91      & \multicolumn{1}{c||}{20.37\textpm0.10}      & 27.64       & 32.34       & 29.81\textpm0.20      \\
                                    & \multicolumn{1}{c||}{LexRank}   & 21.50      & 14.30      & \multicolumn{1}{c||}{16.00\textpm0.50}      & 24.69      & 18.17      & \multicolumn{1}{c||}{20.94\textpm0.10}      & 29.22       & 32.98       & 29.65\textpm0.20      \\
                                    & \multicolumn{1}{c||}{MEAD}      & 29.20      & 27.80      & \multicolumn{1}{c||}{26.80\textpm0.50}      & 26.83      & 28.26      & \multicolumn{1}{c||}{27.52\textpm0.10}      & 25.40       & 41.03       & 31.38\textpm0.10      \\
                                    \hline
                                    & \multicolumn{1}{c||}{SVM}       & 34.30      & 32.70      & \multicolumn{1}{c||}{31.40\textpm0.40}      & 17.09      & 4.32      & \multicolumn{1}{c||}{6.90\textpm0.10}      & 27.19       & 14.09       & 18.56\textpm0.30      \\
                                    & \multicolumn{1}{c||}{LogReg}    & 14.50      & 12.10      & \multicolumn{1}{c||}{12.50\textpm0.50}      & 5.10      & 0.67      & \multicolumn{1}{c||}{1.18\textpm0.30}      & 18.43       & 6.22       & 9.30\textpm0.40      \\ \hline
									& \multicolumn{1}{c||}{LSTM}      & 43.12      & 38.09      & \multicolumn{1}{c||}{40.44\textpm0.03}      & 35.02      & 30.45      & \multicolumn{1}{c||}{32.27\textpm0.02}      & 35.31       & 26.30       & 30.17\textpm0.01      \\
									& \multicolumn{1}{c||}{CNN}       & 35.17      & 23.35      & \multicolumn{1}{c||}{28.03\textpm0.03}      & 27.91      & 22.61      & \multicolumn{1}{c||}{24.98\textpm0.02}      & 27.63       & 26.88       & 26.23\textpm0.01      \\ 
									\hline 
\multicolumn{11}{c}{{\cellcolor[gray]{.9}}Hierarchical + Static Embedding}                                                                                                                                                                        \\ \hline
\multirow{3}{*}{w2v}       			& \multicolumn{1}{c||}{HAN}       & 39.65      & 33.41      & \multicolumn{1}{c||}{\textbf{36.26}\textpm0.05}      & 27.01      & 29.74      & \multicolumn{1}{c||}{28.31\textpm0.02}      & 25.80       & 27.33       & 26.55\textpm0.01      \\
                                    & \multicolumn{1}{c||}{HCNN}      & 36.37      & 26.78      & \multicolumn{1}{c||}{30.84\textpm0.03}      & 26.64      & 23.23      & \multicolumn{1}{c||}{24.82\textpm0.04}      & 23.20       & 31.43       & \textbf{26.69}\textpm0.03      \\
                                    & \multicolumn{1}{c||}{Ours}    & 40.65      & 32.49      & \multicolumn{1}{c||}{36.12\textpm0.02}      & 27.63      & 30.78      & \multicolumn{1}{c||}{\textbf{29.12}\textpm0.04}      & 25.40       & 26.65       & 26.01\textpm0.03      \\ \hline
\multirow{3}{*}{FastText}  			& \multicolumn{1}{c||}{HAN}       & 35.56      & 25.43      & \multicolumn{1}{c||}{29.66\textpm0.05}      & 27.11      & 28.00      & \multicolumn{1}{c||}{\textbf{27.55}\textpm0.03}      & 25.90       & 28.52       & \textbf{27.15}\textpm0.01      \\
                                    & \multicolumn{1}{c||}{HCNN}      & 34.97      & 25.56      & \multicolumn{1}{c||}{29.53\textpm0.03}      & 26.60      & 23.22      & \multicolumn{1}{c||}{24.80\textpm0.03}      & 22.64       & 28.54       & 25.25\textpm0.02      \\
                                    & \multicolumn{1}{c||}{Ours}    & 39.97      & 32.25      & \multicolumn{1}{c||}{\textbf{35.70}\textpm0.02}      & 29.48      & 20.89      & \multicolumn{1}{c||}{24.45\textpm0.03}      & 25.32       & 26.81       & 26.04\textpm0.02      \\ 
                                    \hline
\multicolumn{11}{c}{{\cellcolor[gray]{.9}}Hierarchical + ELMo}                                                                                                                                                                                     \\ \hline
\multirow{3}{*}{}          			& \multicolumn{1}{c||}{HAN}       & 39.74      & 33.48      & \multicolumn{1}{c||}{36.31\textpm0.01}      & 35.60      & 1.98      & \multicolumn{1}{c||}{3.74\textpm0.01}      & 31.61       & 25.84       & 28.44\textpm0.02      \\
                                    & \multicolumn{1}{c||}{HCNN}      & 38.02      & 30.44      & \multicolumn{1}{c||}{33.81\textpm0.01}      & 36.07      & 5.03      & \multicolumn{1}{c||}{\textbf{8.82}\textpm0.01}      & 31.22       & 19.22       & 23.79\textpm0.01      \\
                                    & \multicolumn{1}{c||}{Ours}    & 40.74      & 36.92      & \multicolumn{1}{c||}{\textbf{38.69}\textpm0.02}      & 35.49      & 3.75      & \multicolumn{1}{c||}{6.79\textpm0.01}      & 30.87       & 26.52       & \textbf{28.53}\textpm0.02      \\ \hline
\multirow{3}{*}{+w2v}      			& \multicolumn{1}{c||}{HAN}       & 38.05      & 30.06      & \multicolumn{1}{c||}{33.56\textpm0.01}      & 30.78      & 15.69      & \multicolumn{1}{c||}{\textbf{20.79}\textpm0.01}      & 26.31       & 27.17       & \textbf{26.73}\textpm0.02      \\
                                    & \multicolumn{1}{c||}{HCNN}      & 38.08      & 31.08      & \multicolumn{1}{c||}{34.18\textpm0.01}      & 33.62      & 2.29      & \multicolumn{1}{c||}{4.28\textpm0.01}      & 26.57       & 26.07       & 26.32\textpm0.01      \\
                                    & \multicolumn{1}{c||}{Ours}    & 38.30      & 32.50      & \multicolumn{1}{c||}{\textbf{35.15}\textpm0.01}      & 29.33      & 14.60      & \multicolumn{1}{c||}{19.50\textpm0.02}      & 26.44       & 25.54       & 25.98\textpm0.01      \\ \hline
\multirow{3}{*}{+FastText} 			& \multicolumn{1}{c||}{HAN}       & 37.67      & 29.12      & \multicolumn{1}{c||}{32.83\textpm0.02}      & 30.35      & 15.39      & \multicolumn{1}{c||}{\textbf{20.42}\textpm0.01}      & 25.89       & 24.16       & 24.99\textpm0.01      \\
                                    & \multicolumn{1}{c||}{HCNN}      & 38.20      & 30.30      & \multicolumn{1}{c||}{33.78\textpm0.01}      & 34.45      & 3.56      & \multicolumn{1}{c||}{6.45\textpm0.01}      & 25.79       & 24.22       & 24.98\textpm0.02      \\
                                    & \multicolumn{1}{c||}{Ours}    & 38.16      & 31.88      & \multicolumn{1}{c||}{\textbf{34.72}\textpm0.01}      & 29.61      & 11.63      & \multicolumn{1}{c||}{16.70\textpm0.02}      & 26.80       & 30.66       & \textbf{28.60}\textpm0.02      \\ 
                                    \hline
\multicolumn{11}{c}{{\cellcolor[gray]{.9}}Hierarchical + BERT}                                                                                                                                                                                     \\ \hline
\multirow{3}{*}{}          			& \multicolumn{1}{c||}{HAN}       & 40.94      & 44.00      & \multicolumn{1}{c||}{42.41\textpm0.01}      & 33.96      & 1.14      & \multicolumn{1}{c||}{\textbf{2.19}\textpm0.01}      & 19.71       & 4.40       & \textbf{7.20}\textpm0.01      \\
                                    & \multicolumn{1}{c||}{HCNN}      & 43.63      & 31.60      & \multicolumn{1}{c||}{36.65\textpm0.01}      & 33.76      & 1.13      & \multicolumn{1}{c||}{2.18\textpm0.01}      & 20.09       & 4.39       & \textbf{7.20}\textpm0.01      \\
                                    & \multicolumn{1}{c||}{Ours}    & 41.53      & 46.45      & \multicolumn{1}{c||}{\textbf{43.85}\textpm0.01}      & 33.60      & 1.12      & \multicolumn{1}{c||}{2.16\textpm0.01}      & 19.78       & 4.34       & 7.12\textpm0.01      \\ \hline
\multirow{3}{*}{+w2v}      			& \multicolumn{1}{c||}{HAN}       & 39.69      & 41.53      & \multicolumn{1}{c||}{40.59\textpm0.01}      & 28.15      & 3.32      & \multicolumn{1}{c||}{5.93\textpm0.02}      & 28.15       & 3.32       & 5.93\textpm0.01      \\
                                    & \multicolumn{1}{c||}{HCNN}      & 40.43      & 42.81      & \multicolumn{1}{c||}{\textbf{41.59}\textpm0.01}      & 34.02      & 2.59      & \multicolumn{1}{c||}{4.81\textpm0.01}      & 34.02       & 2.59       & 4.81\textpm0.01      \\
                                    & \multicolumn{1}{c||}{Ours}    & 40.20      & 42.47      & \multicolumn{1}{c||}{41.30\textpm0.01}      & 31.86      & 4.02      & \multicolumn{1}{c||}{\textbf{7.13}\textpm0.03}      & 31.86       & 4.02       & \textbf{7.13}\textpm0.02      \\ \hline
\multirow{3}{*}{+FastText} 			& \multicolumn{1}{c||}{HAN}       & 39.73      & 41.81      & \multicolumn{1}{c||}{40.75\textpm0.01}      & 32.03      & 2.92      & \multicolumn{1}{c||}{\textbf{5.34}\textpm0.02}      & 20.01       & 4.38       & \textbf{7.19}\textpm0.01      \\
                                    & \multicolumn{1}{c||}{HCNN}      & 40.24      & 42.10      & \multicolumn{1}{c||}{41.15\textpm0.02}      & 33.17      & 2.74      & \multicolumn{1}{c||}{5.05\textpm0.02}      & 19.92       & 4.37       & 7.16\textpm0.01      \\
                                    & \multicolumn{1}{c||}{Ours}    & 40.37      & 42.32      & \multicolumn{1}{c||}{\textbf{41.32}\textpm0.01}      & 36.33      & 2.47      & \multicolumn{1}{c||}{4.61\textpm0.01}      & 19.70       & 4.34       & 7.11\textpm0.01     \\
\hline
\end{tabular}
\end{small}
\end{adjustbox}
\label{tab:main_sent_level_results}
\end{table*}

\begin{table*}
\begin{center}
\setlength{\tabcolsep}{8pt}
\caption{Ablation study to investigate the effect of each component in the hierarchical-based models. F1 scores of ROUGE-L are compared (unit in percentage). \Checkmark indicates the component available in the model. The overall improvement in \textcolor{red}{red} and \textcolor{blue}{blue} are the proposed model performance gain compared to Baseline LSTM and Baseline CNN, respectivey.}
\label{tab:component_analysis}
\begin{tabular}{c||c|c|c||c|c|c}
\hline
\multicolumn{1}{c||}{}                                                         & \multicolumn{3}{|c||}{\textbf{Model Component}}                                           & \multicolumn{3}{c}{\textbf{Data}}         \\ 
\cline{2-7}
                                                                             & \textbf{LSTM} & \textbf{CNN} & \begin{tabular}[c]{@{}c@{}}\textbf{Hierarchical}\\ \textbf{Attention}\end{tabular} & \textbf{Trip advisor} & \textbf{Reddit} & \textbf{Newsroom} \\
                                                                             \hline
                                                                             \hline
\multicolumn{1}{c||}{\begin{tabular}[c]{@{}c@{}}Baseline LSTM\end{tabular}} & \Checkmark    &     &                                                                  & 20.07        & 38.81  & 19.87    \\
\multicolumn{1}{c||}{\begin{tabular}[c]{@{}c@{}}Baseline CNN\end{tabular}}  &      & \Checkmark   &                                                                  & 20.07        & 29.23  & 14.82    \\
\multicolumn{1}{c||}{HAN}                                                     & \Checkmark    &     & \Checkmark                                                                & 21.84        & 44.52  & 18.39    \\
\multicolumn{1}{c||}{HCNN}                                                    &      & \Checkmark   & \Checkmark                                                                & 21.60        & 44.61  & 17.25    \\
\multicolumn{1}{c||}{Ours}                                                  & \Checkmark    & \Checkmark   & \Checkmark                                                                & 32.01        & 53.34  & 24.11    \\
\hline
\hline
                                                                             & \multicolumn{3}{c||}{\textbf{Overall Improvement}}                                      & \textcolor{red}{+11.94}        & \textcolor{red}{+14.53}  & \textcolor{red}{+4.24}   \\
                                                                             & \multicolumn{3}{c||}{}                                      & \textcolor{blue}{+11.94}        & \textcolor{blue}{+24.11}  & \textcolor{blue}{+9.29}   \\
\hline
\end{tabular}
\end{center}
\end{table*}
\begin{table*}
\setlength{\tabcolsep}{10pt}
\caption{Summarization results from all models. 
F1 scores are reported in percentage for Rouge-1, Rouge-2, and Rouge-L respectively.}
\begin{adjustbox}{width=\textwidth}
\begin{small}
\begin{tabular}{c||cccccccccc}
\hline
                                    & \multicolumn{1}{c||}{}                   	  & \multicolumn{3}{c||}{\textbf{Trip advisor}}                & \multicolumn{3}{c||}{\textbf{Reddit}}                      & \multicolumn{3}{c}{\textbf{Newsroom}} \\
                                    \cline{3-11}
\textbf{Embedding}		                    & \multicolumn{1}{c||}{\textbf{Method}}    & \textbf{R-1} 		  & \textbf{R-2}		   & \multicolumn{1}{c||}{\textbf{R-L}} 		 & \textbf{R-1} 		  & \textbf{R-2} 	   & \multicolumn{1}{c||}{\textbf{R-L}} 		 & \textbf{R-1}  	   & \textbf{R-2}  		 & \textbf{R-L} \\ 
\hline
\multicolumn{11}{c}{{\cellcolor[gray]{.9}}Baselines} \\ \hline
\multirow{8}{*}{            }       & \multicolumn{1}{c||}{ILP}       & 29.30      & 9.90      & \multicolumn{1}{c||}{12.80}      & 40.85      & 38.96      & \multicolumn{1}{c||}{37.84}      & 17.45       & 16.89       & 16.05      \\
                                    & \multicolumn{1}{c||}{Sum-Basic} & 33.10      & 10.40      & \multicolumn{1}{c||}{13.70}      & 36.66      & 34.72      & \multicolumn{1}{c||}{36.63}      & 16.28       & 15.63       & 16.31      \\
                                    & \multicolumn{1}{c||}{KL-Sum}    & 35.50      & 12.30      & \multicolumn{1}{c||}{13.40}      & 46.87      & 45.23      & \multicolumn{1}{c||}{46.78}      & 21.10       & 20.53       & 20.96      \\
									& \multicolumn{1}{c||}{LSA}   & 34.20      & 14.50      & \multicolumn{1}{c||}{13.60}      & 46.85      & 45.24      & \multicolumn{1}{c||}{46.88}      & 25.29       & 24.80       & 24.41      \\
                                    & \multicolumn{1}{c||}{LexRank}   & 38.70      & 14.20      & \multicolumn{1}{c||}{13.20}      & 44.93      & 43.28      & \multicolumn{1}{c||}{44.85}      & 21.08       & 20.51       & 20.99      \\
                                    & \multicolumn{1}{c||}{MEAD}      & 38.50      & 15.40      & \multicolumn{1}{c||}{22.00}      & 44.70      & 46.94      & \multicolumn{1}{c||}{47.57}      & 20.97       & 22.24       & 22.26      \\
                                    & \multicolumn{1}{c||}{Opinosis}  & 0.62      & 0.10      & \multicolumn{1}{c||}{0.99}      & 1.33      & 0.24      & \multicolumn{1}{c||}{1.67}      & 2.22       & 0.95       & 2.24      \\ 
									& \multicolumn{1}{c||}{TextRank}  & -      & -      & \multicolumn{1}{c||}{-}      & -      & -      & \multicolumn{1}{c||}{-}      & 24.45       & 10.12       & 20.13      \\ 
                                    \hline
                                    & \multicolumn{1}{c||}{SVM}       & 24.70      & 10.00      & \multicolumn{1}{c||}{25.80}      & 6.02      & 2.57      & \multicolumn{1}{c||}{7.46}      & 17.46       & 17.76       & 25.04      \\
                                    & \multicolumn{1}{c||}{LogReg}    & 29.40      & 7.80      & \multicolumn{1}{c||}{10.30}      & 0.73      & 0.34      & \multicolumn{1}{c||}{0.92}      & 7.21       & 7.35       & 11.23      \\  \hline
									& \multicolumn{1}{c||}{LSTM}      & 33.02      & 11.92      & \multicolumn{1}{c||}{20.07}      & 48.00      & 34.40      & \multicolumn{1}{c||}{38.81}      & 24.47       & 15.06       & 19.87      \\
									& \multicolumn{1}{c||}{CNN}       & 33.37      & 12.22      & \multicolumn{1}{c||}{20.07}      & 40.00      & 24.41      & \multicolumn{1}{c||}{29.23}      & 20.46       & 9.19       & 14.82      \\ 
									\hline
\multicolumn{11}{c}{{\cellcolor[gray]{.9}}Hierarchical + Static Embedding}                                                                                                                                                            \\ \hline
\multirow{3}{*}{w2v}       			& \multicolumn{1}{c||}{HAN}       & 36.19      & 13.15      & \multicolumn{1}{c||}{21.84}      & 46.95      & 32.09      & \multicolumn{1}{c||}{44.52}      & 24.26       & 12.30       & 18.39      \\
                                    & \multicolumn{1}{c||}{HCNN}      & 36.34      & 12.74      & \multicolumn{1}{c||}{21.60}      & 46.71      & 31.63      & \multicolumn{1}{c||}{44.61}      & 23.68       & 11.35       & 17.25      \\
                                    & \multicolumn{1}{c||}{Ours}    & \textbf{38.13}      & \textbf{15.51}      & \multicolumn{1}{c||}{\textbf{32.01}}      & \textbf{54.67}      & \textbf{42.84}      & \multicolumn{1}{c||}{\textbf{53.34}}      & \textbf{25.56}       & \textbf{12.41}       & \textbf{24.11}      \\ \hline
\multirow{3}{*}{FastText}  			& \multicolumn{1}{c||}{HAN}       & 37.23      & 13.29      & \multicolumn{1}{c||}{22.05}      & 46.62      & 31.73      & \multicolumn{1}{c||}{44.17}      & 24.38       & \textbf{12.54}       & 18.46      \\
                                    & \multicolumn{1}{c||}{HCNN}      & 36.55      & 13.20      & \multicolumn{1}{c||}{22.29}      & 43.82      & 27.31      & \multicolumn{1}{c||}{41.07}      & 23.66       & 11.23       & 17.29      \\
                                    & \multicolumn{1}{c||}{Ours}    & \textbf{37.10}      & \textbf{14.02}      & \multicolumn{1}{c||}{\textbf{30.31}}      & \textbf{53.86}      & \textbf{41.83}      & \multicolumn{1}{c||}{\textbf{52.55}}      & \textbf{25.33}       & 12.03       & \textbf{23.81}      \\ 
                                    \hline
\multicolumn{11}{c}{{\cellcolor[gray]{.9}}Hierarchical + ELMo}                                                                                                                                                       				  \\ \hline
\multirow{3}{*}{}          			& \multicolumn{1}{c||}{HAN}       & 35.86      & 13.94      & \multicolumn{1}{c||}{30.00}      & 41.62      & 26.00      & \multicolumn{1}{c||}{30.57}      & 25.01       & 12.88       & 18.88      \\
                                    & \multicolumn{1}{c||}{HCNN}      & \textbf{36.29}      & \textbf{14.34}      & \multicolumn{1}{c||}{\textbf{30.75}}      & 41.88      & 26.40      & \multicolumn{1}{c||}{30.73}      & 25.14       & 12.95       & 18.53      \\
                                    & \multicolumn{1}{c||}{Ours}    & 35.67      & 13.36      & \multicolumn{1}{c||}{30.21}      & \textbf{44.88}      & \textbf{28.96}      & \multicolumn{1}{c||}{\textbf{42.16}}      & \textbf{26.45}       & \textbf{13.59}       & \textbf{24.99}      \\ \hline
\multirow{3}{*}{+w2v}      			& \multicolumn{1}{c||}{HAN}       & 35.80      & 13.83      & \multicolumn{1}{c||}{28.62}      & 40.60      & 24.82      & \multicolumn{1}{c||}{30.40}      & 24.32       & 12.06       & 17.69      \\
                                    & \multicolumn{1}{c||}{HCNN}      & 35.90      & 13.78      & \multicolumn{1}{c||}{\textbf{29.23}}      & 40.85      & 25.72      & \multicolumn{1}{c||}{30.95}      & 23.77       & 11.60       & 17.60      \\
                                    & \multicolumn{1}{c||}{Ours}    & \textbf{35.92}      & \textbf{14.07}      & \multicolumn{1}{c||}{28.85}      & \textbf{43.48}      & \textbf{27.28}      & \multicolumn{1}{c||}{\textbf{41.20}}      & \textbf{25.87}       & \textbf{12.77}       & \textbf{24.43}      \\ \hline
\multirow{3}{*}{+FastText} 			& \multicolumn{1}{c||}{HAN}       & 36.26      & 14.43      & \multicolumn{1}{c||}{30.05}      & 41.68      & 25.52      & \multicolumn{1}{c||}{30.46}      & 23.43       & 11.38       & 17.35      \\
                                    & \multicolumn{1}{c||}{HCNN}      & 35.81      & 13.88      & \multicolumn{1}{c||}{30.41}      & 41.63      & 26.27      & \multicolumn{1}{c||}{30.85}      & 23.67       & 11.64       & 17.21      \\
                                    & \multicolumn{1}{c||}{Ours}    & \textbf{37.17}      & \textbf{15.19}      & \multicolumn{1}{c||}{\textbf{30.64}}      & \textbf{44.17}      & \textbf{27.90}      & \multicolumn{1}{c||}{\textbf{41.48}}      & \textbf{25.69}       & \textbf{12.54}       & \textbf{24.27}      \\ 
                                    \hline
\multicolumn{11}{c}{{\cellcolor[gray]{.9}}Hierarchical + BERT}                                                                                                                                                       				  \\ \hline
\multirow{3}{*}{}          			& \multicolumn{1}{c||}{HAN}       & 35.02      & 10.84      & \multicolumn{1}{c||}{19.50}      & 40.24      & 23.64      & \multicolumn{1}{c||}{28.78}      & 25.88       & 14.23       & 19.69      \\
                                    & \multicolumn{1}{c||}{HCNN}      & 34.78      & 10.84      & \multicolumn{1}{c||}{19.71}      & 40.08      & 23.42      & \multicolumn{1}{c||}{28.51}      & 26.06       & 14.37       & 19.83      \\
                                    & \multicolumn{1}{c||}{Ours}    & \textbf{36.13}      & \textbf{11.96}      & \multicolumn{1}{c||}{\textbf{29.28}}      & \textbf{42.08}      & \textbf{25.23}      & \multicolumn{1}{c||}{\textbf{39.63}}      & \textbf{27.51}       & \textbf{15.33}       & \textbf{26.58}      \\ \hline
\multirow{3}{*}{+w2v}      			& \multicolumn{1}{c||}{HAN}       & 30.58      & 9.24      & \multicolumn{1}{c||}{19.15}      & 40.65      & 24.50      & \multicolumn{1}{c||}{29.20}      & 25.93       & 14.29       & 19.51      \\
                                    & \multicolumn{1}{c||}{HCNN}      & 30.72      & 9.25      & \multicolumn{1}{c||}{18.75}      & 40.52      & 24.30      & \multicolumn{1}{c||}{28.89}      & 25.56       & 13.92       & 19.58      \\
                                    & \multicolumn{1}{c||}{Ours}    & \textbf{33.21}      & \textbf{11.24}      & \multicolumn{1}{c||}{\textbf{29.19}}      & \textbf{42.51}      & \textbf{26.15}      & \multicolumn{1}{c||}{\textbf{40.21}}      & \textbf{26.16}       & \textbf{14.44}       & \textbf{20.08}      \\ \hline
\multirow{3}{*}{+FastText} 			& \multicolumn{1}{c||}{HAN}       & 33.32      & 10.24      & \multicolumn{1}{c||}{19.29}      & 45.47      & 31.13      & \multicolumn{1}{c||}{35.55}      & 25.64       & 14.34       & 19.76      \\
                                    & \multicolumn{1}{c||}{HCNN}      & 34.04      & 10.65      & \multicolumn{1}{c||}{19.44}      & 45.17      & 30.55      & \multicolumn{1}{c||}{35.18}      & 25.34       & 13.91       & 19.31      \\
                                    & \multicolumn{1}{c||}{Ours}    & \textbf{35.14}      & \textbf{12.03}      & \multicolumn{1}{c||}{\textbf{29.23}}      & \textbf{47.66}      & \textbf{32.88}      & \multicolumn{1}{c||}{\textbf{45.48}}      & \textbf{27.43}       & \textbf{15.25}       & \textbf{26.45}     \\
\hline
\end{tabular}
\end{small}
\end{adjustbox}
\label{tab:main_summ_results}
\end{table*}

\subsection{Ablation Study}
In this subsection, we discuss our observations from extensive ablation experiments conducted to better understand our model from various aspects. 

\subsubsection{Model Component Analysis}
A comprehensive component analysis is performed by adding different components on top of baseline methods as presented in Table \ref{tab:component_analysis}. The results reveal that using a model equipped with either CNN or LSTM alone performs poorly across all datasets. This indicates that leveraging the hierarchical structure of an input document to generate a document representation has helped boost the performance. Specifically, the hierarchical structure captures information at both word and sentence levels -- word-level representation is learned and subsequently aggregated to form a sentence; likewise, sentence-level representation is learned and subsequently aggregated to form document representation. Among all hierarchical-based models, the ROUGE-L scores of HAN and HCNN are comparative, whereas our proposed model has illustrated an evident performance gain. The shift in improvement is also noticeable ($>$1\%)  for Reddit and Newsroom datasets which are larger in size than Trip advisor.

According to Table \ref{tab:component_analysis}, when comparing results from the proposed model against those from baseline-CNN and baseline-LSTM, the overall improvement is 11.94\% for Trip advisor dataset, 14.53\% and 24.11\% for Reddit dataset, and 4.24\% and 9.29\% for Newsroom dataset, respectively.

\subsubsection{Effect of CNN configurations}
Table \ref{tab:breadth_analysis} and \ref{tab:depth_analysis} show the model performance on different receptive field sizes and different number of convolutional layers, while other parameters remain fixed. It is important to note that when multiple receptive field sizes are used, such as [2,3], an output obtained from each local feature needs to be concatenated first to yield a representation that will be used by a layer following CNN. We observe that the receptive field size of 2 outperforms others across all datasets in both sentence classification and ROUGE evaluation. For Reddit dataset, the decrease in performance is notable. Regarding the number of convolutional layers, we observe that increasing convolutional layers leads to a drop in overall performance. With the number of layers of 6 (highest), the classification performance as well as the output summary quality are the lowest. The CNN part of our proposed model, therefore, applies a receptive field size of 2 and a single convolutional layer.    

\begin{table*}
\begin{center}
\setlength{\tabcolsep}{8pt}
\caption{Ablation study to investigate the effect of receptive field size towards the overall performance improvement. F1 scores are reported for both sentence-level classification (SL) and ROUGE evaluation (R-1, R-2, R-L).
Shaded in gray are best values (in \%). Non-shaded values presents loss compared to the best values, also in \%.}
\label{tab:breadth_analysis}
\begin{tabular}{c||c|c|c|c||c|c|c|c||c|c|c|c}
\hline
              &    \multicolumn{4}{c||}{\textbf{Trip advisor}} &     \multicolumn{4}{c||}{\textbf{Reddit}} &    \multicolumn{4}{c}{\textbf{Newsroom}} \\
              \cline{2-13}
\textbf{Size} & \textbf{SL} & \textbf{R-1}       & \textbf{R-2}       & \textbf{R-L}       & \textbf{SL} & \textbf{R-1}     & \textbf{R-2}     & \textbf{R-L}     & \textbf{SL} & \textbf{R-1}      & \textbf{R-2}     & \textbf{R-L}     \\ \hline
\hline
\rowcolor[gray]{.9}
\textbf{2}     			& 36.12 & 38.13     & 15.51     & 32.01     & 29.12 & 54.67   & 42.84   & 53.34   & 26.01 & 25.56    & 12.41   & 24.11   \\
\textbf{{[}2,3{]}}     	& -2.41 & -2.44     & -1.45     & -1.65     & -8.33 & -10.51   & -14.72   & -11.68   & -2.00 & -2.90    & -1.69   & -2.32   \\
\textbf{{[}2,3,4{]}}     & -2.64 & -1.97     & -0.97     & -1.23     & -12.33 & -11.03   & -15.06   & -11.86   & -1.15 & -2.76    & -1.56   & -2.24   \\
\textbf{{[}2,3,4,5{]}}   & -2.91 & -2.37     & -1.44     & -1.60     & -7.84 & -10.56   & -14.71   & -11.55   & -0.66 & -3.17    & -1.88   & -2.48   \\
\hline
\end{tabular}
\end{center}
\end{table*}
\begin{table*}
\begin{center}
\setlength{\tabcolsep}{8pt}
\caption{Ablation study to investigate the effect of number of convolutional layer(s) towards the overall performance improvement. F1 scores are reported for both sentence-level classification (SL) and ROUGE evaluation (R-1, R-2, R-L). Shaded in gray are best values (in \%). Non-shaded values presents loss compared to the best values, also in \%.}
\label{tab:depth_analysis}
\begin{tabular}{c||c|c|c|c||c|c|c|c||c|c|c|c}
\hline
          & \multicolumn{4}{c||}{\textbf{Trip advisor}} &    \multicolumn{4}{c||}{\textbf{Reddit}} &     \multicolumn{4}{c}{\textbf{Newsroom}} \\
          \cline{2-13}
\textbf{Depth} & \textbf{SL} & \textbf{R-1}       & \textbf{R-2}       & \textbf{R-L}       & \textbf{SL} & \textbf{R-1}     & \textbf{R-2}     & \textbf{R-L}     & \textbf{SL} & \textbf{R-1}      & \textbf{R-2}     & \textbf{R-L}     \\ 
\hline
\hline
\rowcolor[gray]{.9}
\textbf{1}     & 36.12 & 38.13     & 15.51     & 32.01     & 29.12 & 54.67   & 42.84   & 53.34   & 26.01 & 25.56    & 12.41   & 24.11   \\
\textbf{2}     & -2.02 & -2.11     & -1.11     & -1.17     & -2.01 & -11.07   & -15.59   & -12.17   & -0.97 & -2.98    & -1.72   & -2.49   \\
\textbf{3}     & -1.28 & -2.36     & -1.39     & -1.57     & -10.33 & -10.68   & -14.69   & -11.68   & -2.30 & -3.15    & -2.13   & -2.70   \\
\textbf{4}     & -2.03 & -2.30     & -1.10     & -1.42     & -12.04 & -10.86   & -14.87   & -11.72   & -1.64 & -3.04    & -1.87   & -2.39   \\
\textbf{5}     & -1.80 & -2.48     & -1.51     & -1.75     & -9.97 & -10.82   & -15.08   & -11.80   & -0.94 & -2.84    & -1.66   & -2.31   \\
\textbf{6}     & -2.34 & -2.14     & -1.10     & -1.15     & -10.55 & -11.27   & -15.39   & -12.23   & -3.07 & -3.37    & -2.29   & -2.80    \\
\hline
\end{tabular}
\end{center}
\end{table*}

\subsubsection{Representation Learning}
In this section, we discuss the impact of different embeddings on the hierarchical models. We report the effect of using static word embeddings versus contextual embeddings. We also concatenate static and contextual embeddings to examine their joint effect on the performance.

We investigate the outputs from the model initialized with only static word embeddings. The results for all datasets in Table \ref{tab:main_sent_level_results} show that using word2vec mostly is superior to FastText in classifying sentence labels as well as in terms of ROUGE evaluation. With respect to contextual representations, for Trip advisor dataset, the results show that BERT embeddings yield better classification performance in terms of sentence-level scores than ELMo. For the remaining datasets, however, the sentence-level results mostly reveal a significant drop in the recall and F1-scores as a consequence. As aforementioned in Section \ref{sec:compare_hier}, this indicates that the system determines few sentences as summary-worthy; yet, most of these sentences are correctly labelled. From Table \ref{tab:main_summ_results}, the ROUGE evaluation shows that, in general using static word embeddings achieves better performance.

In addition, inspired by the study of \cite{peters-etal-2018-deep}, we concatenated static and contextual representations at the \textit{sentence-level encoders}. In terms of sentence-level scores, the results show that the concatenation does not significantly affect the performance, except for ELMo+ w2v and ELMo+FastText in Reddit dataset, of which a significant improvement can be noticed in HAN and the proposed models. With respect to ROUGE evaluation, the results obtained from concatenated representation also do not reflect a significant improvement.

\begin{table}
\begin{center}
\setlength{\tabcolsep}{2pt}
\caption{Ablation study to investigate the effect of attention mechanism. The results are obtained from the proposed model. \Checkmark means the attention mechanism is applied in the model, whereas \XSolidBrush is the opposite case.}

\label{tab:attention_analysis}
\begin{tabular}{c||cc||cc||cc||cc}
\hline
             & \multicolumn{2}{c||}{\textbf{Sentence-level}} & \multicolumn{2}{c||}{\textbf{R-1}} & \multicolumn{2}{c||}{\textbf{R-2}} & \multicolumn{2}{c}{\textbf{R-L}} \\
\textbf{Data}         & \Checkmark                &  \XSolidBrush               & \Checkmark           & \XSolidBrush          & \Checkmark           & \XSolidBrush          & \Checkmark           & \XSolidBrush          \\ \hline \hline
\textbf{Trip advisor} & 36.12            & \textbf{36.13}            & \textbf{38.13}       & 35.74      & \textbf{15.51}       & 14.4      & \textbf{32.01}       & 30.75      \\
\textbf{Reddit}       & 29.12            & \textbf{30.33}            & \textbf{54.67}       & 43.55      & \textbf{42.84}       & 27.30      & \textbf{53.34}       & 41.00      \\
\textbf{Newsroom}     & \textbf{26.01}            & 24.14            & \textbf{25.56}       & 19.73      & \textbf{12.41}       & 9.20      & \textbf{24.11}       & 14.65     \\
\hline
\end{tabular}
\end{center}
\end{table}
\subsubsection{Effect of Attention Mechanism on Selecting Salient Units}
We investigate the attention layer to validate whether the attention mechanism aids in selecting representative units. Table \ref{tab:attention_analysis} shows that, with respect to ROUGE evaluation, when the attention mechanism is incorporated in the model, the performance is improved across all datasets. In particular, for the larger dataset (Reddit and Newsroom), the difference of results between with and without attention is nontrivial. In terms of sentence-level classification, incorporating the attention mechanism does not significantly affect the performance.

It is important to note that the proposed model attends to important \textit{bigrams} at the word level and contiguous \textit{sentence pairs} at the sentence level. At the word level, the attention value of each bigram influences the sentence vector to which the bigram belongs. The attention weight is computed according to the relevance of each bigram, given the sentence context. If a sentence contains many bigrams with high attention values, its corresponding sentence vector will potentially contain information about these prominent bigrams. In the sentence level, likewise, the attention values of the sentence pairs influence the resulting thread vector. A high attention value of a sentence pair indicates its importance and relevance towards the thread key concept. This attention-weighted sentence pair goes through softmax normalization, from which the output indicates how likely a sentence pair is a key unit for the summary.

\begin{figure*}
\centering
    \begin{subfigure}{0.8\textwidth}
    \centering\includegraphics[width=\linewidth]{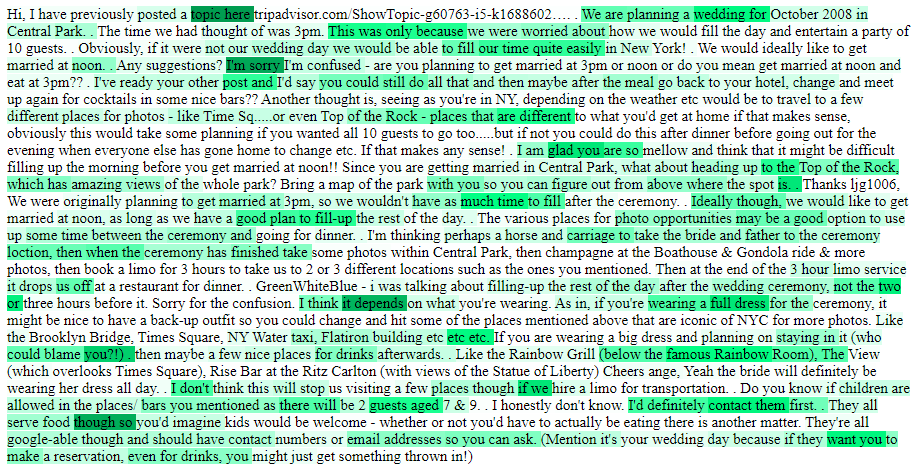}

  \end{subfigure}
    \begin{subfigure}{0.75\textwidth}
    \centering\includegraphics[width=\linewidth]{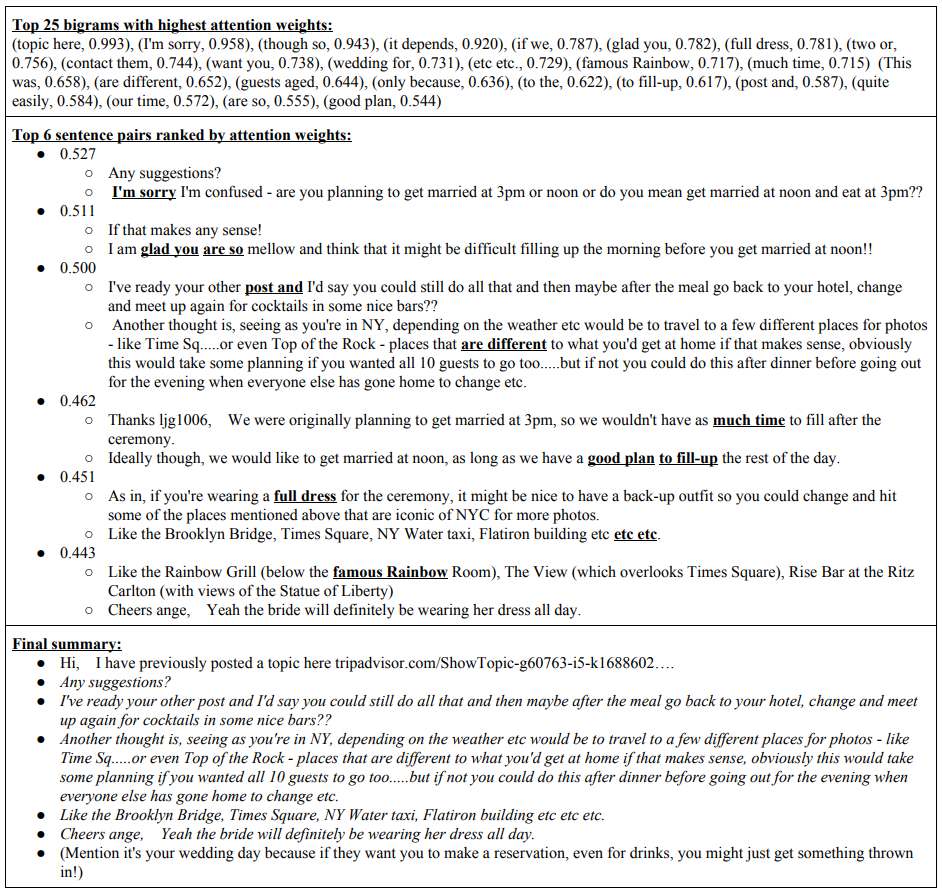}
  \end{subfigure}\\
  \caption{Visualization of the generated summary for a forum thread. Top: Relative attention weights for each bigram by the proposed model over an entire thread. Bigrams with a darker highlight present higher importance. The attention values of all bigrams were obtained from word-level attention layer.
  Bottom: The first row presents a list of top 25 bigrams that are ranked according to their attention values, formatted as a (bigram, attention weights) tuple.  The second row presents a list of sentence pairs ranked by attention weights, where the highest weight is 0.527. The bigrams in bold and underlined are those with highest attention weights. The third row presents the final summary which lists all chronologically-ordered extracted sentences. The sentences in italic are those in the top 6 sentence pairs with highest attention weights.}
\label{fig:word_attention}
\end{figure*}

Figure \ref{fig:word_attention} illustrates a visualization of words in the example summary. The bigram with high attention weight will be highlighted with a darker shade compared to other bigrams with lower attention. The sentence \textit{“I am glad you are so mellow and think that it might be difficult filling up the morning before you get married at noon!!”} contains two bigrams, namely \textit{“glad you”} and \textit{“are so”} which have attention weights of 0.782 and 0.555, respectively. The sentence encoder outputs a weighted sum of the bigrams using normalized attention as weight, and the two aforementioned bigrams are represented the most in the encoded sentence. Later in the thread encoder, this sentence has also shown to be in one of the highest sentence pairs ranked by attention weights. Finally, in the final summary, it can be noticed that the majority of sentences (italicized) are those belonging to the example sentence pairs with high ranked weights. Nevertheless, we emphasize that it is \textit{not} necessarily the case that if a sentence is in a sentence pair with high attention, it will be selected into the final summary. High attention weights only indicate the significance of the constituent unit. In other words, whether or not a sentence is chosen into the summary is determined by the output layer which considers both sentence and thread representations concatenated together. However, it is observed that when sentences belong to sentence pairs with high attention weights, they have a higher chance of being selected into the final thread summary.


\section{Conclusions}\label{sec:conclusion}
In this study, we present Hierarchical Unified Deep Neural Network to extractively summarize online forum threads. Our proposed network utilizes two deep neural networks, namely Bi-LSTM and CNN, to obtain representations used to classify whether or not the sentence is summary-worthy. 
The experimental results on three real-life datasets have demonstrated that the proposed model outperforms the majority of baseline methods. 

Our findings confirm the initial hypothesis that the capability of encoders can be enhanced through the proposed architecture. In essence, Bi-LSTM serves a role to capture contextual information, whereas CNN helps to signify prominent units that are keys pertaining to a summary. Together, the strength of both deep neural networks have been leveraged to achieve effective representations. 

Finally, we have conducted extensive experiments to investigate the effect of attention mechanism and pretrained embeddings. The results show that applying attention to the high-level features extracted and compressed by CNN, together with the contextual embeddings, provides a promising avenue towards improving the performance of extractive summarization methods. 

\ifCLASSOPTIONcompsoc
  \section*{Acknowledgments}
\else
  \section*{Acknowledgment}
\fi

This work was supported by Crystal Photonics, Inc. (Grant 1063271).




%



\bibliographystyle{IEEEtran}
\bibliography{manuscript_HybridNetwork}

\begin{thebibliography}{10}
\providecommand{\url}[1]{#1}
\csname url@samestyle\endcsname
\providecommand{\newblock}{\relax}
\providecommand{\bibinfo}[2]{#2}
\providecommand{\BIBentrySTDinterwordspacing}{\spaceskip=0pt\relax}
\providecommand{\BIBentryALTinterwordstretchfactor}{4}
\providecommand{\BIBentryALTinterwordspacing}{\spaceskip=\fontdimen2\font plus
\BIBentryALTinterwordstretchfactor\fontdimen3\font minus
  \fontdimen4\font\relax}
\providecommand{\BIBforeignlanguage}[2]{{%
\expandafter\ifx\csname l@#1\endcsname\relax
\typeout{** WARNING: IEEEtran.bst: No hyphenation pattern has been}%
\typeout{** loaded for the language `#1'. Using the pattern for}%
\typeout{** the default language instead.}%
\else
\language=\csname l@#1\endcsname
\fi
#2}}
\providecommand{\BIBdecl}{\relax}
\BIBdecl

\bibitem{cheng2016neural}
\BIBentryALTinterwordspacing
J.~Cheng and M.~Lapata, ``Neural summarization by extracting sentences and
  words,'' in \emph{Proceedings of the 54th Annual Meeting of the Association
  for Computational Linguistics (Volume 1: Long Papers)}.\hskip 1em plus 0.5em
  minus 0.4em\relax Berlin, Germany: Association for Computational Linguistics,
  Aug. 2016, pp. 484--494. [Online]. Available:
  \url{https://www.aclweb.org/anthology/P16-1046}
\BIBentrySTDinterwordspacing

\bibitem{yang-etal-2016-hierarchical}
\BIBentryALTinterwordspacing
Z.~Yang, D.~Yang, C.~Dyer, X.~He, A.~Smola, and E.~Hovy, ``Hierarchical
  attention networks for document classification,'' in \emph{Proceedings of the
  2016 Conference of the North {A}merican Chapter of the Association for
  Computational Linguistics: Human Language Technologies}.\hskip 1em plus 0.5em
  minus 0.4em\relax San Diego, California: Association for Computational
  Linguistics, Jun. 2016, pp. 1480--1489. [Online]. Available:
  \url{https://www.aclweb.org/anthology/N16-1174}
\BIBentrySTDinterwordspacing

\bibitem{zhou-etal-2018-neural-document}
\BIBentryALTinterwordspacing
Q.~Zhou, N.~Yang, F.~Wei, S.~Huang, M.~Zhou, and T.~Zhao, ``Neural document
  summarization by jointly learning to score and select sentences,'' in
  \emph{Proceedings of the 56th Annual Meeting of the Association for
  Computational Linguistics (Volume 1: Long Papers)}.\hskip 1em plus 0.5em
  minus 0.4em\relax Melbourne, Australia: Association for Computational
  Linguistics, Jul. 2018, pp. 654--663. [Online]. Available:
  \url{https://www.aclweb.org/anthology/P18-1061}
\BIBentrySTDinterwordspacing

\bibitem{tarnpradab2017toward}
S.~Tarnpradab, F.~Liu, and K.~A. Hua, ``Toward extractive summarization of
  online forum discussions via hierarchical attention networks,'' in \emph{The
  Thirtieth International Flairs Conference}, 2017.

\bibitem{hahn2000challenges}
U.~Hahn and I.~Mani, ``The challenges of automatic summarization,''
  \emph{Computer}, vol.~33, no.~11, pp. 29--36, 2000.

\bibitem{bhatia-etal-2014-summarizing}
\BIBentryALTinterwordspacing
S.~Bhatia, P.~Biyani, and P.~Mitra, ``Summarizing online forum discussions {--}
  can dialog acts of individual messages help?'' in \emph{Proceedings of the
  2014 Conference on Empirical Methods in Natural Language Processing
  ({EMNLP})}.\hskip 1em plus 0.5em minus 0.4em\relax Doha, Qatar: Association
  for Computational Linguistics, Oct. 2014, pp. 2127--2131. [Online].
  Available: \url{https://www.aclweb.org/anthology/D14-1226}
\BIBentrySTDinterwordspacing

\bibitem{hu2017opinion}
Y.-H. Hu, Y.-L. Chen, and H.-L. Chou, ``Opinion mining from online hotel
  reviews--a text summarization approach,'' \emph{Information Processing \&
  Management}, vol.~53, no.~2, pp. 436--449, 2017.

\bibitem{10.1145/3357384.3358161}
\BIBentryALTinterwordspacing
H.~Liu and X.~Wan, ``Neural review summarization leveraging user and product
  information,'' in \emph{Proceedings of the 28th ACM International Conference
  on Information and Knowledge Management}, ser. CIKM ’19.\hskip 1em plus
  0.5em minus 0.4em\relax New York, NY, USA: Association for Computing
  Machinery, 2019, p. 2389–2392. [Online]. Available:
  \url{https://doi.org/10.1145/3357384.3358161}
\BIBentrySTDinterwordspacing

\bibitem{10.1145/3368926.3369699}
\BIBentryALTinterwordspacing
M.-T. Nguyen, T.~V. Cuong, and N.~X. Hoai, ``Exploiting user comments for
  document summarization with matrix factorization,'' in \emph{Proceedings of
  the Tenth International Symposium on Information and Communication
  Technology}, ser. SoICT 2019.\hskip 1em plus 0.5em minus 0.4em\relax New
  York, NY, USA: Association for Computing Machinery, 2019, p. 118–124.
  [Online]. Available: \url{https://doi.org/10.1145/3368926.3369699}
\BIBentrySTDinterwordspacing

\bibitem{di2017surf}
A.~Di~Sorbo, S.~Panichella, C.~V. Alexandru, C.~A. Visaggio, and G.~Canfora,
  ``Surf: summarizer of user reviews feedback,'' in \emph{2017 IEEE/ACM 39th
  International Conference on Software Engineering Companion (ICSE-C)}.\hskip
  1em plus 0.5em minus 0.4em\relax IEEE, 2017, pp. 55--58.

\bibitem{carenini-etal-2008-summarizing}
\BIBentryALTinterwordspacing
G.~Carenini, R.~T. Ng, and X.~Zhou, ``Summarizing emails with conversational
  cohesion and subjectivity,'' in \emph{Proceedings of ACL-08: HLT}.\hskip 1em
  plus 0.5em minus 0.4em\relax Columbus, Ohio: Association for Computational
  Linguistics, Jun. 2008, pp. 353--361. [Online]. Available:
  \url{https://www.aclweb.org/anthology/P08-1041}
\BIBentrySTDinterwordspacing

\bibitem{10.1145/3274465}
\BIBentryALTinterwordspacing
A.~X. Zhang and J.~Cranshaw, ``Making sense of group chat through collaborative
  tagging and summarization,'' \emph{Proc. ACM Hum.-Comput. Interact.}, vol.~2,
  no. CSCW, Nov. 2018. [Online]. Available:
  \url{https://doi.org/10.1145/3274465}
\BIBentrySTDinterwordspacing

\bibitem{10.1145/3159652.3160588}
\BIBentryALTinterwordspacing
N.~Tepper, A.~Hashavit, M.~Barnea, I.~Ronen, and L.~Leiba, ``Collabot:
  Personalized group chat summarization,'' in \emph{Proceedings of the Eleventh
  ACM International Conference on Web Search and Data Mining}, ser. WSDM
  ’18.\hskip 1em plus 0.5em minus 0.4em\relax New York, NY, USA: Association
  for Computing Machinery, 2018, p. 771–774. [Online]. Available:
  \url{https://doi.org/10.1145/3159652.3160588}
\BIBentrySTDinterwordspacing

\bibitem{10.1145/3279981.3279987}
\BIBentryALTinterwordspacing
F.~Nihei, Y.~I. Nakano, and Y.~Takase, ``Fusing verbal and nonverbal
  information for extractive meeting summarization,'' in \emph{Proceedings of
  the Group Interaction Frontiers in Technology}, ser. GIFT’18.\hskip 1em
  plus 0.5em minus 0.4em\relax New York, NY, USA: Association for Computing
  Machinery, 2018. [Online]. Available:
  \url{https://doi.org/10.1145/3279981.3279987}
\BIBentrySTDinterwordspacing

\bibitem{10.1145/2993148.2993160}
\BIBentryALTinterwordspacing
------, ``Meeting extracts for discussion summarization based on multimodal
  nonverbal information,'' in \emph{Proceedings of the 18th ACM International
  Conference on Multimodal Interaction}, ser. ICMI ’16.\hskip 1em plus 0.5em
  minus 0.4em\relax New York, NY, USA: Association for Computing Machinery,
  2016, p. 185–192. [Online]. Available:
  \url{https://doi.org/10.1145/2993148.2993160}
\BIBentrySTDinterwordspacing

\bibitem{rudra2018identifying}
K.~Rudra, P.~Goyal, N.~Ganguly, P.~Mitra, and M.~Imran, ``Identifying
  sub-events and summarizing disaster-related information from microblogs,'' in
  \emph{The 41st International ACM SIGIR Conference on Research \& Development
  in Information Retrieval}, 2018, pp. 265--274.

\bibitem{10.1145/3178541}
\BIBentryALTinterwordspacing
K.~Rudra, N.~Ganguly, P.~Goyal, and S.~Ghosh, ``Extracting and summarizing
  situational information from the twitter social media during disasters,''
  \emph{ACM Trans. Web}, vol.~12, no.~3, Jul. 2018. [Online]. Available:
  \url{https://doi.org/10.1145/3178541}
\BIBentrySTDinterwordspacing

\bibitem{10.1145/3357384.3358020}
\BIBentryALTinterwordspacing
A.~Sharma, K.~Rudra, and N.~Ganguly, ``Going beyond content richness: Verified
  information aware summarization of crisis-related microblogs,'' in
  \emph{Proceedings of the 28th ACM International Conference on Information and
  Knowledge Management}, ser. CIKM ’19.\hskip 1em plus 0.5em minus
  0.4em\relax New York, NY, USA: Association for Computing Machinery, 2019, p.
  921–930. [Online]. Available: \url{https://doi.org/10.1145/3357384.3358020}
\BIBentrySTDinterwordspacing

\bibitem{zhou2016cminer}
X.~Zhou, X.~Wan, and J.~Xiao, ``Cminer: opinion extraction and summarization
  for chinese microblogs,'' \emph{IEEE Transactions on Knowledge and Data
  Engineering}, vol.~28, no.~7, pp. 1650--1663, 2016.

\bibitem{10.1145/3377407}
\BIBentryALTinterwordspacing
S.-H. Liu, K.-Y. Chen, and B.~Chen, ``Enhanced language modeling with proximity
  and sentence relatedness information for extractive broadcast news
  summarization,'' \emph{ACM Trans. Asian Low-Resour. Lang. Inf. Process.},
  vol.~19, no.~3, Feb. 2020. [Online]. Available:
  \url{https://doi.org/10.1145/3377407}
\BIBentrySTDinterwordspacing

\bibitem{10.1145/2980258.2980442}
\BIBentryALTinterwordspacing
J.~R. Thomas, S.~K. Bharti, and K.~S. Babu, ``Automatic keyword extraction for
  text summarization in e-newspapers,'' in \emph{Proceedings of the
  International Conference on Informatics and Analytics}, ser. ICIA-16.\hskip
  1em plus 0.5em minus 0.4em\relax New York, NY, USA: Association for Computing
  Machinery, 2016. [Online]. Available:
  \url{https://doi.org/10.1145/2980258.2980442}
\BIBentrySTDinterwordspacing

\bibitem{10.1145/3289600.3291008}
\BIBentryALTinterwordspacing
Y.~Duan and A.~Jatowt, ``Across-time comparative summarization of news
  articles,'' in \emph{Proceedings of the Twelfth ACM International Conference
  on Web Search and Data Mining}, ser. WSDM ’19.\hskip 1em plus 0.5em minus
  0.4em\relax New York, NY, USA: Association for Computing Machinery, 2019, p.
  735–743. [Online]. Available: \url{https://doi.org/10.1145/3289600.3291008}
\BIBentrySTDinterwordspacing

\bibitem{sethi2017automatic}
P.~Sethi, S.~Sonawane, S.~Khanwalker, and R.~Keskar, ``Automatic text
  summarization of news articles,'' in \emph{2017 International Conference on
  Big Data, IoT and Data Science (BID)}.\hskip 1em plus 0.5em minus 0.4em\relax
  IEEE, 2017, pp. 23--29.

\bibitem{nallapati2017summarunner}
R.~Nallapati, F.~Zhai, and B.~Zhou, ``Summarunner: A recurrent neural network
  based sequence model for extractive summarization of documents,'' in
  \emph{Proceedings of the Thirty-First AAAI Conference on Artificial
  Intelligence}, ser. AAAI’17.\hskip 1em plus 0.5em minus 0.4em\relax AAAI
  Press, 2017, p. 3075–3081.

\bibitem{cao2017improving}
Z.~Cao, W.~Li, S.~Li, and F.~Wei, ``Improving multi-document summarization via
  text classification,'' in \emph{Proceedings of the Thirty-First AAAI
  Conference on Artificial Intelligence}, ser. AAAI’17.\hskip 1em plus 0.5em
  minus 0.4em\relax AAAI Press, 2017, p. 3053–3059.

\bibitem{10.1145/3132847.3133127}
\BIBentryALTinterwordspacing
A.~K. Singh, M.~Gupta, and V.~Varma, ``Hybrid memnet for extractive
  summarization,'' in \emph{Proceedings of the 2017 ACM on Conference on
  Information and Knowledge Management}, ser. CIKM ’17.\hskip 1em plus 0.5em
  minus 0.4em\relax New York, NY, USA: Association for Computing Machinery,
  2017, p. 2303–2306. [Online]. Available:
  \url{https://doi.org/10.1145/3132847.3133127}
\BIBentrySTDinterwordspacing

\bibitem{narayan-etal-2018-ranking}
\BIBentryALTinterwordspacing
S.~Narayan, S.~B. Cohen, and M.~Lapata, ``Ranking sentences for extractive
  summarization with reinforcement learning,'' in \emph{Proceedings of the 2018
  Conference of the North {A}merican Chapter of the Association for
  Computational Linguistics: Human Language Technologies, Volume 1 (Long
  Papers)}.\hskip 1em plus 0.5em minus 0.4em\relax New Orleans, Louisiana:
  Association for Computational Linguistics, Jun. 2018, pp. 1747--1759.
  [Online]. Available: \url{https://www.aclweb.org/anthology/N18-1158}
\BIBentrySTDinterwordspacing

\bibitem{10.1145/3308558.3313619}
\BIBentryALTinterwordspacing
Z.~Zhao, H.~Pan, C.~Fan, Y.~Liu, L.~Li, M.~Yang, and D.~Cai, ``Abstractive
  meeting summarization via hierarchical adaptive segmental network learning,''
  in \emph{The World Wide Web Conference}, ser. WWW ’19.\hskip 1em plus 0.5em
  minus 0.4em\relax New York, NY, USA: Association for Computing Machinery,
  2019, p. 3455–3461. [Online]. Available:
  \url{https://doi.org/10.1145/3308558.3313619}
\BIBentrySTDinterwordspacing

\bibitem{10.1145/3308558.3313707}
\BIBentryALTinterwordspacing
J.-Y. Jiang, M.~Zhang, C.~Li, M.~Bendersky, N.~Golbandi, and M.~Najork,
  ``Semantic text matching for long-form documents,'' in \emph{The World Wide
  Web Conference}, ser. WWW ’19.\hskip 1em plus 0.5em minus 0.4em\relax New
  York, NY, USA: Association for Computing Machinery, 2019, p. 795–806.
  [Online]. Available: \url{https://doi.org/10.1145/3308558.3313707}
\BIBentrySTDinterwordspacing

\bibitem{10.1145/3341105.3374025}
\BIBentryALTinterwordspacing
A.~F. Cruz, G.~Rocha, and H.~L. Cardoso, ``On document representations for
  detection of biased news articles,'' in \emph{Proceedings of the 35th Annual
  ACM Symposium on Applied Computing}, ser. SAC ’20.\hskip 1em plus 0.5em
  minus 0.4em\relax New York, NY, USA: Association for Computing Machinery,
  2020, p. 892–899. [Online]. Available:
  \url{https://doi.org/10.1145/3341105.3374025}
\BIBentrySTDinterwordspacing

\bibitem{rush-etal-2015-neural}
\BIBentryALTinterwordspacing
A.~M. Rush, S.~Chopra, and J.~Weston, ``A neural attention model for
  abstractive sentence summarization,'' in \emph{Proceedings of the 2015
  Conference on Empirical Methods in Natural Language Processing}.\hskip 1em
  plus 0.5em minus 0.4em\relax Lisbon, Portugal: Association for Computational
  Linguistics, Sep. 2015, pp. 379--389. [Online]. Available:
  \url{https://www.aclweb.org/anthology/D15-1044}
\BIBentrySTDinterwordspacing

\bibitem{10.1145/3341105.3373892}
\BIBentryALTinterwordspacing
H.~Lee, Y.~Choi, and J.-H. Lee, ``Attention history-based attention for
  abstractive text summarization,'' in \emph{Proceedings of the 35th Annual ACM
  Symposium on Applied Computing}, ser. SAC ’20.\hskip 1em plus 0.5em minus
  0.4em\relax New York, NY, USA: Association for Computing Machinery, 2020, p.
  1075–1081. [Online]. Available:
  \url{https://doi.org/10.1145/3341105.3373892}
\BIBentrySTDinterwordspacing

\bibitem{nema2017diversity}
P.~Nema, M.~Khapra, A.~Laha, and B.~Ravindran, ``Diversity driven attention
  model for query-based abstractive summarization,'' \emph{arXiv preprint
  arXiv:1704.08300}, 2017.

\bibitem{wang-ling-2016-neural}
\BIBentryALTinterwordspacing
L.~Wang and W.~Ling, ``Neural network-based abstract generation for opinions
  and arguments,'' in \emph{Proceedings of the 2016 Conference of the North
  {A}merican Chapter of the Association for Computational Linguistics: Human
  Language Technologies}.\hskip 1em plus 0.5em minus 0.4em\relax San Diego,
  California: Association for Computational Linguistics, Jun. 2016, pp. 47--57.
  [Online]. Available: \url{https://www.aclweb.org/anthology/N16-1007}
\BIBentrySTDinterwordspacing

\bibitem{cao-etal-2016-attsum}
\BIBentryALTinterwordspacing
Z.~Cao, W.~Li, S.~Li, F.~Wei, and Y.~Li, ``{A}tt{S}um: Joint learning of
  focusing and summarization with neural attention,'' in \emph{Proceedings of
  {COLING} 2016, the 26th International Conference on Computational
  Linguistics: Technical Papers}.\hskip 1em plus 0.5em minus 0.4em\relax Osaka,
  Japan: The COLING 2016 Organizing Committee, Dec. 2016, pp. 547--556.
  [Online]. Available: \url{https://www.aclweb.org/anthology/C16-1053}
\BIBentrySTDinterwordspacing

\bibitem{narayan-etal-2018-document}
\BIBentryALTinterwordspacing
S.~Narayan, R.~Cardenas, N.~Papasarantopoulos, S.~B. Cohen, M.~Lapata, J.~Yu,
  and Y.~Chang, ``Document modeling with external attention for sentence
  extraction,'' in \emph{Proceedings of the 56th Annual Meeting of the
  Association for Computational Linguistics (Volume 1: Long Papers)}.\hskip 1em
  plus 0.5em minus 0.4em\relax Melbourne, Australia: Association for
  Computational Linguistics, Jul. 2018, pp. 2020--2030. [Online]. Available:
  \url{https://www.aclweb.org/anthology/P18-1188}
\BIBentrySTDinterwordspacing

\bibitem{10.1145/3269206.3269251}
\BIBentryALTinterwordspacing
C.~Feng, F.~Cai, H.~Chen, and M.~de~Rijke, ``Attentive encoder-based extractive
  text summarization,'' in \emph{Proceedings of the 27th ACM International
  Conference on Information and Knowledge Management}, ser. CIKM ’18.\hskip
  1em plus 0.5em minus 0.4em\relax New York, NY, USA: Association for Computing
  Machinery, 2018, p. 1499–1502. [Online]. Available:
  \url{https://doi.org/10.1145/3269206.3269251}
\BIBentrySTDinterwordspacing

\bibitem{camacho2018word}
J.~Camacho-Collados and M.~T. Pilehvar, ``From word to sense embeddings: A
  survey on vector representations of meaning,'' \emph{Journal of Artificial
  Intelligence Research}, vol.~63, pp. 743--788, 2018.

\bibitem{mikolov2013distributed}
T.~Mikolov, I.~Sutskever, K.~Chen, G.~Corrado, and J.~Dean, ``Distributed
  representations of words and phrases and their compositionality,'' in
  \emph{Proceedings of the 26th International Conference on Neural Information
  Processing Systems - Volume 2}, ser. NIPS’13.\hskip 1em plus 0.5em minus
  0.4em\relax Red Hook, NY, USA: Curran Associates Inc., 2013, p. 3111–3119.

\bibitem{pennington-etal-2014-glove}
\BIBentryALTinterwordspacing
J.~Pennington, R.~Socher, and C.~Manning, ``{G}love: Global vectors for word
  representation,'' in \emph{Proceedings of the 2014 Conference on Empirical
  Methods in Natural Language Processing ({EMNLP})}.\hskip 1em plus 0.5em minus
  0.4em\relax Doha, Qatar: Association for Computational Linguistics, Oct.
  2014, pp. 1532--1543. [Online]. Available:
  \url{https://www.aclweb.org/anthology/D14-1162}
\BIBentrySTDinterwordspacing

\bibitem{bojanowski2017enriching}
P.~Bojanowski, E.~Grave, A.~Joulin, and T.~Mikolov, ``Enriching word vectors
  with subword information,'' \emph{Transactions of the Association for
  Computational Linguistics}, vol.~5, pp. 135--146, 2017.

\bibitem{peters-etal-2018-deep}
\BIBentryALTinterwordspacing
M.~Peters, M.~Neumann, M.~Iyyer, M.~Gardner, C.~Clark, K.~Lee, and
  L.~Zettlemoyer, ``Deep contextualized word representations,'' in
  \emph{Proceedings of the 2018 Conference of the North {A}merican Chapter of
  the Association for Computational Linguistics: Human Language Technologies,
  Volume 1 (Long Papers)}.\hskip 1em plus 0.5em minus 0.4em\relax New Orleans,
  Louisiana: Association for Computational Linguistics, Jun. 2018, pp.
  2227--2237. [Online]. Available:
  \url{https://www.aclweb.org/anthology/N18-1202}
\BIBentrySTDinterwordspacing

\bibitem{devlin-etal-2019-bert}
\BIBentryALTinterwordspacing
J.~Devlin, M.-W. Chang, K.~Lee, and K.~Toutanova, ``{BERT}: Pre-training of
  deep bidirectional transformers for language understanding,'' in
  \emph{Proceedings of the 2019 Conference of the North {A}merican Chapter of
  the Association for Computational Linguistics: Human Language Technologies,
  Volume 1 (Long and Short Papers)}.\hskip 1em plus 0.5em minus 0.4em\relax
  Minneapolis, Minnesota: Association for Computational Linguistics, Jun. 2019,
  pp. 4171--4186. [Online]. Available:
  \url{https://www.aclweb.org/anthology/N19-1423}
\BIBentrySTDinterwordspacing

\bibitem{akbik-etal-2018-contextual}
\BIBentryALTinterwordspacing
A.~Akbik, D.~Blythe, and R.~Vollgraf, ``Contextual string embeddings for
  sequence labeling,'' in \emph{Proceedings of the 27th International
  Conference on Computational Linguistics}.\hskip 1em plus 0.5em minus
  0.4em\relax Santa Fe, New Mexico, USA: Association for Computational
  Linguistics, Aug. 2018, pp. 1638--1649. [Online]. Available:
  \url{https://www.aclweb.org/anthology/C18-1139}
\BIBentrySTDinterwordspacing

\bibitem{hochreiter1997long}
S.~Hochreiter and J.~Schmidhuber, ``Long short-term memory,'' \emph{Neural
  computation}, vol.~9, no.~8, pp. 1735--1780, 1997.

\bibitem{wubben2015facilitating}
S.~Wubben, S.~Verberne, E.~Krahmer, and A.~van~den Bosch, ``Facilitating online
  discussions by automatic summarization,'' 2015.

\bibitem{grusky2018newsroom}
M.~Grusky, M.~Naaman, and Y.~Artzi, ``Newsroom: A dataset of 1.3 million
  summaries with diverse extractive strategies,'' \emph{arXiv preprint
  arXiv:1804.11283}, 2018.

\bibitem{ruder2016overview}
S.~Ruder, ``An overview of gradient descent optimization algorithms,''
  \emph{arXiv preprint arXiv:1609.04747}, 2016.

\bibitem{hinton2012neural}
G.~Hinton, N.~Srivastava, and K.~Swersky, ``Neural networks for machine
  learning lecture 6a overview of mini-batch gradient descent.''

\bibitem{DBLP:journals/corr/abs-1212-5701}
\BIBentryALTinterwordspacing
M.~D. Zeiler, ``{ADADELTA:} an adaptive learning rate method,'' \emph{CoRR},
  vol. abs/1212.5701, 2012. [Online]. Available:
  \url{http://arxiv.org/abs/1212.5701}
\BIBentrySTDinterwordspacing

\bibitem{duchi2011adaptive}
J.~Duchi, E.~Hazan, and Y.~Singer, ``Adaptive subgradient methods for online
  learning and stochastic optimization,'' \emph{Journal of machine learning
  research}, vol.~12, no. Jul, pp. 2121--2159, 2011.

\bibitem{kingma2014adam}
D.~P. Kingma and J.~Ba, ``Adam: A method for stochastic optimization,''
  \emph{arXiv preprint arXiv:1412.6980}, 2014.

\bibitem{lin-2004-rouge}
\BIBentryALTinterwordspacing
C.-Y. Lin, ``{ROUGE}: A package for automatic evaluation of summaries,'' in
  \emph{Text Summarization Branches Out}.\hskip 1em plus 0.5em minus
  0.4em\relax Barcelona, Spain: Association for Computational Linguistics, Jul.
  2004, pp. 74--81. [Online]. Available:
  \url{https://www.aclweb.org/anthology/W04-1013}
\BIBentrySTDinterwordspacing

\bibitem{ganesan2015rouge}
K.~Ganesan, ``Rouge 2.0: Updated and improved measures for evaluation of
  summarization tasks,'' 2015.

\bibitem{zhang-wallace-2017-sensitivity}
\BIBentryALTinterwordspacing
Y.~Zhang and B.~Wallace, ``A sensitivity analysis of (and practitioners{'}
  guide to) convolutional neural networks for sentence classification,'' in
  \emph{Proceedings of the Eighth International Joint Conference on Natural
  Language Processing (Volume 1: Long Papers)}.\hskip 1em plus 0.5em minus
  0.4em\relax Taipei, Taiwan: Asian Federation of Natural Language Processing,
  Nov. 2017, pp. 253--263. [Online]. Available:
  \url{https://www.aclweb.org/anthology/I17-1026}
\BIBentrySTDinterwordspacing

\bibitem{mikolov2018advances}
T.~Mikolov, E.~Grave, P.~Bojanowski, C.~Puhrsch, and A.~Joulin, ``Advances in
  pre-training distributed word representations,'' in \emph{Proceedings of the
  International Conference on Language Resources and Evaluation (LREC 2018)},
  2018.

\bibitem{berg-kirkpatrick-etal-2011-jointly}
\BIBentryALTinterwordspacing
T.~Berg-Kirkpatrick, D.~Gillick, and D.~Klein, ``Jointly learning to extract
  and compress,'' in \emph{Proceedings of the 49th Annual Meeting of the
  Association for Computational Linguistics: Human Language
  Technologies}.\hskip 1em plus 0.5em minus 0.4em\relax Portland, Oregon, USA:
  Association for Computational Linguistics, Jun. 2011, pp. 481--490. [Online].
  Available: \url{https://www.aclweb.org/anthology/P11-1049}
\BIBentrySTDinterwordspacing

\bibitem{boudin-etal-2015-concept}
\BIBentryALTinterwordspacing
F.~Boudin, H.~Mougard, and B.~Favre, ``Concept-based summarization using
  integer linear programming: From concept pruning to multiple optimal
  solutions,'' in \emph{Proceedings of the 2015 Conference on Empirical Methods
  in Natural Language Processing}.\hskip 1em plus 0.5em minus 0.4em\relax
  Lisbon, Portugal: Association for Computational Linguistics, Sep. 2015, pp.
  1914--1918. [Online]. Available:
  \url{https://www.aclweb.org/anthology/D15-1220}
\BIBentrySTDinterwordspacing

\bibitem{Vanderwende:2007}
L.~Vanderwende, H.~Suzuki, C.~Brockett, and A.~Nenkova, ``Beyond {SumBasic}:
  {T}ask-focused summarization with sentence simplification and lexical
  expansion,'' \emph{Information Processing and Management}, vol.~43, no.~6,
  pp. 1606--1618, 2007.

\bibitem{haghighi-vanderwende-2009-exploring}
\BIBentryALTinterwordspacing
A.~Haghighi and L.~Vanderwende, ``Exploring content models for multi-document
  summarization,'' in \emph{Proceedings of Human Language Technologies: The
  2009 Annual Conference of the North {A}merican Chapter of the Association for
  Computational Linguistics}.\hskip 1em plus 0.5em minus 0.4em\relax Boulder,
  Colorado: Association for Computational Linguistics, Jun. 2009, pp. 362--370.
  [Online]. Available: \url{https://www.aclweb.org/anthology/N09-1041}
\BIBentrySTDinterwordspacing

\bibitem{steinberger2004using}
J.~Steinberger \emph{et~al.}, ``Using latent semantic analysis in text
  summarization and summary evaluation.''

\bibitem{Erkan:2004}
G.~Erkan and D.~R. Radev, ``Lexrank: Graph-based lexical centrality as salience
  in text summarization,'' \emph{J. Artif. Int. Res.}, vol.~22, no.~1, p.
  457–479, Dec. 2004.

\bibitem{Radev:2004}
D.~R. Radev, H.~Jing, M.~Sty\'{s}, and D.~Tam, ``Centroid-based summarization
  of multiple documents,'' \emph{Information Processing and Management},
  vol.~40, no.~6, pp. 919--938, 2004.

\bibitem{ganesan2010opinosis}
K.~Ganesan, C.~Zhai, and J.~Han, ``Opinosis: a graph-based approach to
  abstractive summarization of highly redundant opinions,'' in
  \emph{Proceedings of the 23rd International Conference on Computational
  Linguistics}.\hskip 1em plus 0.5em minus 0.4em\relax Association for
  Computational Linguistics, 2010, pp. 340--348.

\bibitem{barrios2016variations}
F.~Barrios, F.~L{\'o}pez, L.~Argerich, and R.~Wachenchauzer, ``Variations of
  the similarity function of textrank for automated summarization,''
  \emph{arXiv preprint arXiv:1602.03606}, 2016.

\bibitem{Fan:2008}
R.-E. Fan, K.-W. Chang, C.-J. Hsieh, X.-R. Wang, and C.-J. Lin, ``{LIBLINEAR}:
  {A} library for large linear classification,'' \emph{Journal of Machine
  Learning Research}, vol.~9, pp. 1871--1874, 2008.

\bibitem{kim-2014-convolutional}
\BIBentryALTinterwordspacing
Y.~Kim, ``Convolutional neural networks for sentence classification,'' in
  \emph{Proceedings of the 2014 Conference on Empirical Methods in Natural
  Language Processing ({EMNLP})}.\hskip 1em plus 0.5em minus 0.4em\relax Doha,
  Qatar: Association for Computational Linguistics, Oct. 2014, pp. 1746--1751.
  [Online]. Available: \url{https://www.aclweb.org/anthology/D14-1181}
\BIBentrySTDinterwordspacing

\bibitem{Dang:2008}
H.~T. Dang and K.~Owczarzak, ``Overview of the {TAC} 2008 update summarization
  task,'' in \emph{Proceedings of Text Analysis Conference (TAC)}, 2008.

\bibitem{luo-etal-2016-automatic}
\BIBentryALTinterwordspacing
W.~Luo, F.~Liu, Z.~Liu, and D.~Litman, ``Automatic summarization of student
  course feedback,'' in \emph{Proceedings of the 2016 Conference of the North
  {A}merican Chapter of the Association for Computational Linguistics: Human
  Language Technologies}.\hskip 1em plus 0.5em minus 0.4em\relax San Diego,
  California: Association for Computational Linguistics, Jun. 2016, pp. 80--85.
  [Online]. Available: \url{https://www.aclweb.org/anthology/N16-1010}
\BIBentrySTDinterwordspacing

\end{thebibliography}

%

\begin{IEEEbiographynophoto}{Sansiri Tarnpradab (member, IEEE)}
received her B.Eng. degree in Computer Engineering from King Mongkut's University of Technology Thonburi, Bangkok, Thailand in 2011 and M.S. degree in Computer Engineering from Syracuse University, USA in 2013. She is currently working towards the Ph.D. degree in the Department of Computer Science, University of Central Florida. Her research interests include Social Media Analytics, Natural Language Processing, and Deep Learning. Her current focus is Deep Recurrent Neural Networks for Social Media challenges.
\end{IEEEbiographynophoto}

\begin{IEEEbiographynophoto}{Fereshteh Jafariakinabad (member, IEEE)}
recieved the B.Sc. degree from Amirkabir University of Technology, Tehran, Iran. She is currently a PhD candidate at University of Central Florida. Her research interest includes Natural Language Processing, more specifically Style-based text classification and authorship attribution.
\end{IEEEbiographynophoto}

\begin{IEEEbiographynophoto}{Kien A. Hua (Fellow, IEEE)}
is a Pegasus Professor and Director
of the Data Systems Lab at the University
of Central Florida. He was the Associate
Dean for Research of the College of Engineering
and Computer Science at UCF. Prior to joining
the university, he was a Lead Architect at IBM
Mid-Hudson Laboratory, where he led a team
of senior engineers to develop a highly parallel
computer system, the precursor to the highly
successful commercial parallel computer known
as SP2. Dr. Hua received his B.S. in Computer
Science, and M.S. and Ph.D. in Electrical Engineering, all from the
University of Illinois at Urbana-Champaign, USA. His diverse expertise
includes multimedia computing, machine learning, Internet of Things,
network and wireless communications, and mobile computing. He has
published widely with 13 papers recognized as best/top papers at conferences
and a journal. Dr. Hua has served as a Conference Chair,
an Associate Chair, and a Technical Program Committee Member of
numerous international conferences, and on the editorial boards of
several professional journals. Dr. Hua is a Fellow of IEEE.
\end{IEEEbiographynophoto}







\end{document}